\documentclass[10pt,twocolumn,letterpaper]{article}

\usepackage{cvpr}
\usepackage{times}
\usepackage{epsfig}
\usepackage{graphicx}
\usepackage{amsmath}
\usepackage{amssymb}
\usepackage{lipsum}
\usepackage{authblk}
\usepackage{fancyhdr}

\usepackage[usestackEOL]{stackengine}

\usepackage{times}
\usepackage{epsfig}
\usepackage{graphicx}
\usepackage{float}
\usepackage{wrapfig}
\usepackage{amsmath,amssymb,amsthm}
\usepackage{algorithm,algorithmicx,algpseudocode}
\usepackage{bm,xspace}
\usepackage{comment}
\usepackage{verbatim}
\usepackage{multirow}
\usepackage{balance}
\usepackage{url}
\usepackage{booktabs}
\usepackage{etoolbox,siunitx}
\usepackage{calc}
\usepackage{pifont,hologo}
\usepackage[usenames, dvipsnames]{color}
\usepackage{nicefrac}

\usepackage[super]{nth}
\usepackage{nicefrac}
\robustify\bfseries

\newcommand*\samethanks[1][\value{footnote}]{\footnotemark[#1]}

\def\gG{\mathcal{G}}

\def\rR{\mathcal{R}}

\def\deg{^{\circ}}

\DeclareMathSymbol{@}{\mathord}{letters}{"3B}

\newcommand\abs[1]{\left\lvert#1\right\rvert}
\newcommand\norm[1]{\left\lVert#1\right\rVert}

\definecolor{dblue}{rgb}{0,0,0.7}

\newcommand\mypara[1]{\vspace{1mm}\noindent\textbf{#1}}

\def\rotc#1{\rotatebox[origin=c]{90}{#1}}

\def\latex/{\LaTeX}
\def\bibtex/{\hologo{BibTeX}}

\usepackage{subfig}
\usepackage[font={small}]{caption}

\usepackage{xcolor}
\usepackage[percent]{overpic}

\usepackage[pagebackref=true,breaklinks=true,letterpaper=true,colorlinks,bookmarks=false]{hyperref}

\fancypagestyle{firstpage}{%
  \fancyhf{}%
  \fancyfoot[C]{\fontsize{8}{10} \selectfont \textcopyright 2019 IEEE.  Personal use of this material is permitted.  Permission from IEEE must be obtained for all other uses, in any current or future media, including reprinting/republishing this material for advertising or promotional purposes, creating new collective works, for resale or redistribution to servers or lists, or reuse of any copyrighted component of this work in other works.}%
}

\cvprfinalcopy

\def\cvprPaperID{2029}

\ifcvprfinal\fi
\begin{document}

\title{What Do Single-view 3D Reconstruction Networks Learn?}

\author[1]{Maxim Tatarchenko\thanks{Equal contribution.}}
\author[2]{Stephan R. Richter\samethanks{}}
\author[2]{Ren\'e Ranftl}
\author[2]{Zhuwen Li}
\author[2]{\\Vladlen Koltun}
\author[1]{Thomas Brox}

\affil{University of Freiburg \enspace \enspace \enspace $^2$Intel Labs}

\makeatletter
\g@addto@macro\@maketitle{
	\begin{figure}[H]
		\setlength{\linewidth}{\textwidth}
		\setlength{\hsize}{\textwidth}
		\vspace{-7mm}
		\centering
		\begin{overpic}[width=17.5cm]{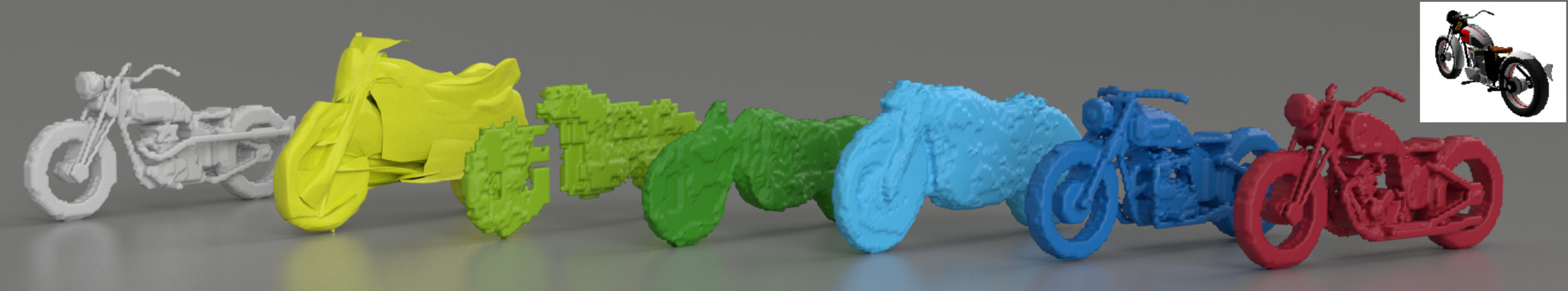}
 		\end{overpic}
		\vspace{-1.6em}
		\caption{We provide evidence that state-of-the-art single-view 3D reconstruction methods (AtlasNet (light green, $0.38$ IoU) \cite{groueix18}, OGN (green, \mbox{$0.46$ IoU) \cite{tatarchenko17}}, Matryoshka Networks (dark green, $0.47$ IoU) \cite{richter18}) do not actually perform reconstruction but image classification. We explicitly design pure recognition baselines (Clustering (light blue, $0.46$ IoU) and Retrieval (dark blue, $0.57$ IoU)) and show that they produce similar or better results both qualitatively and quantitatively. For reference, we show the ground truth (white) and a nearest neighbor from the training set (red, $0.76$ IoU). The inset shows the input image.}
		\label{fig:teaser}
	\end{figure}
}

\makeatother
\maketitle
\thispagestyle{firstpage}

\begin{abstract}
   Convolutional networks for single-view object reconstruction have shown impressive performance and have become a popular subject of research. All existing techniques are united by the idea of having an encoder-decoder network that performs non-trivial reasoning about the 3D structure of the output space. In this work, we set up two alternative approaches that perform image classification and retrieval respectively. These simple baselines yield better results than state-of-the-art methods, both qualitatively and quantitatively. We show that encoder-decoder methods are statistically indistinguishable from these baselines, thus indicating that the current state of the art in single-view object reconstruction does not actually perform reconstruction but image classification. We identify aspects of popular experimental procedures that elicit this behavior and discuss ways to improve the current state of research.
\end{abstract}

\section{Introduction}
\label{sec:introduction}
Object-based single-view 3D reconstruction calls for generating the 3D model of an object given a single image.
Consider the motorcycle in Fig.~\ref{fig:teaser}.
Inferring its 3D structure requires a complex process that combines low-level image cues, knowledge about structural arrangement of parts, and high-level semantic information. We refer to the extremes of this spectrum as \textit{reconstruction} and \textit{recognition}.
\textit{Reconstruction} implies reasoning about the 3D structure of the input image using cues such as texture, shading, and perspective effects. 
\textit{Recognition} amounts to classifying the input image and retrieving the most suitable 3D model from a database, in our example finding a pre-existing 3D model of a motorcycle based on the input image.

While various architectures and 3D representations have been proposed in the literature, existing methods for single-view 3D understanding all use an encoder-decoder structure, where the encoder maps the input image to a latent representation and the decoder is supposed to perform non-trivial reasoning about the 3D structure of the output space. To solve the task, the overall network is expected to incorporate low-level as well as high-level information.

In this work, we analyze the results of state-of-the-art encoder-decoder methods \cite{groueix18, richter18, tatarchenko17} and find that they rely primarily on recognition to address the single-view 3D reconstruction task, while showing only limited reconstruction abilities.
To support this claim, we design two pure recognition baselines: one that combines 3D shape clustering and image classification and one that performs image-based 3D shape retrieval.
Based on these, we demonstrate that the performance of modern convolutional networks for single-view 3D reconstruction can be surpassed even without explicitly inferring the 3D structure of objects.
In many cases the predictions of the recognition baselines are not only better quantitatively, but also appear visually more appealing, as demonstrated in Fig.~\ref{fig:teaser}.

We argue that the dominance of recognition in convolutional networks for single-view 3D reconstruction is a consequence of certain aspects of popular experimental procedures, including dataset composition and evaluation protocols. These allow the network to find a shortcut solution, which happens to be image recognition.

\section{Related work}
\label{sec:related_work}
Historically, single-image 3D reconstruction has been approached via shape-from-shading \cite{durou08,Horn70,zhang99}. More exotic cues for reconstruction are texture \cite{Loh06} and defocus \cite{FS05}. These techniques only reason about visible parts of a surface using a single depth cue. More general approaches for depth estimation from a single monocular image use multiple cues as well as structural knowledge to infer an estimate of the depth of visible surfaces. Saxena \etal \cite{saxena05} estimated depth from a single image by training an MRF on local and global image features. Oswald \etal. \cite{oswald12} solved the same problem with interactive user input. Hoiem \etal \cite{Hoiem05} used recognition together with simple geometric assumptions to construct 3D models from a single image. Karsch \etal \cite{karsch14} proposed a non-parametric framework that uses part- and object-level recognition to assemble an estimate from a database of images and corresponding depth maps. More recently, significant advances have been made in monocular depth estimation by employing convolutional networks \cite{Chen2016, eigen14, Godard2017, LiSnavely2018, Xian2018}.

This paper focuses on methods that not only reason about the 3D structure of object parts visible in the input image, but also hallucinate the invisible parts using priors learned from data.
Tulsiani \etal~\cite{tulsiani15} approached this task with deformable models for specific object categories. 
Most of the recent methods trained convolutional networks that map 2D images to 3D shapes using direct 3D supervision.
A cluster of approaches used voxel-based representations of 3D shapes and generated them with 3D up-convolutions from a latent representation~\cite{choy16, girdhar16, wu16}.
Several works~\cite{haene17, riegler17, tatarchenko17} performed hierarchical partitioning of the output space to achieve computational and memory efficiency, which allows predicting higher-resolution 3D shapes.
Johnston \etal~\cite{johnston17} reconstructed high-resolution 3D shapes with an inverse discrete cosine transform decoder.
Wang \etal~\cite{wang18} generated meshes by deforming a sphere into a desired shape, assuming a fixed distance between camera and objects.
Groueix \etal~\cite{groueix18} assembled surfaces from small patches.
Multiple methods~\cite{lin18, lun17, soltani17, tatarchenko16} produced multi-view depth maps that are fused together into an output point cloud.
Richter \etal~\cite{richter18} extended this with nested shapes fused into a single voxel grid.
Fan \etal~\cite{fan17} directly regressed point clouds.
Wu \etal~\cite{wu17} learned the mapping from input images to 2.5D sketches in a fully-supervised fashion, and then trained a network to map these intermediate representations to the final 3D shapes.
Kong \etal~\cite{Kong17} use 2D landmark locations together with silhouettes to retrieve and deform CAD models. Pontes \etal~\cite{Pontes17} improved upon this work by using a free-form deformation parametrization to model shape variation.

Tulsiani \etal \cite{tulsiani17abstraction} and Niu \etal \cite{niu18} aimed for structural 3D understanding, approximating 3D shapes with a pre-defined set of primitives.

Recently, there has been a trend towards using weaker forms of supervision for single-view 3D shape prediction with convolutional networks.
Multiple approaches \cite{kato18, rezende16, tulsiani17, yan16, zhu17} trained shape regressors by comparing projections of ground-truth and predicted shapes.
Kanazawa \etal~\cite{kanazawa18} predicted deformations from mean shapes trained from multiple learning signals.

There are only very few datasets available for the task of single-image 3D reconstruction -- a consequence of the cost of data collection.
Most existing methods use subsets of ShapeNet~\cite{shapenet15} for training and testing.
Recently, Wiles and Zisserman~\cite{wiles17} introduced two new synthetic datasets: Blobby objects and Sculptures.
The Pix3D dataset~\cite{sun18} provides pairs of perfectly aligned natural images and CAD models.
This dataset, however, contains a low number of 3D samples, which is problematic for training deep networks.

\section{Reconstruction vs. recognition}
\label{sec:the_three_rs}
Single-view 3D understanding is a complex task that requires interpreting visual data both geometrically and semantically. In fact, these two modes are not disjoint, but span a spectrum from pure geometric reconstruction to pure semantic recognition.

\mypara{Reconstruction} implies per-pixel reasoning about the 3D structure of the object shown in the input image, which can be achieved by using low-level image cues such as color, texture, shading, perspective, shadows, and defocus. This mode does not require semantic understanding of the image content.

\mypara{Recognition} is an extreme case of using semantic priors: it operates on the level of whole objects and amounts to classifying the object in the input image and retrieving a corresponding 3D shape from a database.
While it provides a robust prior for reasoning about the invisible parts of objects, this kind of purely semantic solution is only valid if the new object can be explained by an object in the database.

As reconstruction and recognition represent opposing ends of a spectrum, resorting exclusively to either is unlikely to produce the most accurate 3D shapes, since both ignore valuable information present in the input image. It is thus commonly hypothesized that a successful approach to single-view 3D reconstruction needs to combine low-level image cues, structural knowledge, and high-level object understanding~\cite{saxena07}.

In the following sections, we argue that current methods tackle the problem predominantly using recognition.

\section{Conventional setup}
\label{sec:conventional_evaluation}
In this section, we analyze current methods for single-view 3D reconstruction and their relation to reconstruction and recognition.
We employ a standard setup for single-view 3D shape estimation.
We use the ShapeNet dataset~\cite{shapenet15}.
Unlike several recent approaches, which evaluated only on the 13 largest classes,
we deliberately use all 55 classes, as was done in \cite{shapenet17iccv}.
This allows us to investigate how the number of samples within a class influences shape estimation performance.
Within each class, the shapes are randomly split into training, validation, and test sets, containing 70\%, 10\%, and 20\% of the samples respectively.
Every shape was rendered using the ShapeNet-Viewer from five uniformly sampled viewpoints $\left(\theta_{azimuth} \in [0\deg, 360\deg), \theta_{elevation} \in [0\deg, 50\deg)\right)$. The distance to the camera was set such that each rendered shape roughly fits the frame. 
We rendered RGB images of size $224\times 224$, which were downsampled to the input resolution that is required by each method.

All 3D shapes have a consistent canonical orientation and are represented as $128^3$ voxel grids. Using high-resolution ground truth (compared to the conventionally used $32^3$ voxel grids) is crucial for evaluating a method's ability to reconstruct fine detail. Evaluating on a higher resolution than $128^3$ does not offer additional benefits, since the performance of state-of-the-art methods saturates at this level \cite{richter18, tatarchenko17}, while training and evaluation become much more costly.
We follow standard procedure and measure shape similarity with the mean Intersection over Union (mIoU) metric, aggregating predictions within semantic classes \cite{choy16, fan17, haene17, richter18, shin18, tatarchenko17, yan16}.

\subsection{Existing approaches}
We base our experiments on modern convolutional networks that predict high-resolution 3D models from a single image. A taxonomy of approaches arises by categorizing them based on their output representation: voxel grids, meshes, point clouds, and depth maps.
To this end, we chose state-of-the-art methods that cover the dominant output representations or have clearly shown to outperform other related representations for our evaluation.

We use Octree Generating Networks (OGN)~\cite{tatarchenko17} as the representative method that predicts the output directly on a voxel grid. Compared to earlier works \cite{choy16} that operate on this output representation, OGN allows predicting higher-resolution shapes by using octrees to represent the occupied space efficiently. 
We evaluate AtlasNet \cite{groueix18} as the representative approach for surface-based methods. AtlasNet predicts a collection of parametric surfaces and constitutes the state-of-the-art among methods that operate on this output representation. It was shown to outperform the only approach that directly produces point clouds as output \cite{fan17}, as well as another octree-based approach \cite{haene17}. 
Finally, we evaluate the current state-of-the-art in the field, Matryoshka Networks~\cite{richter18}. Matryoshka Networks use a shape representation that is composed of multiple, nested depth maps, which are volumetrically fused into a single output object.

For IoU-based evaluation of the surface predictions from AtlasNet, we project them to depth maps, which we further fuse to a volumetric representation. In our experiments, this approach reliably closed holes in the reconstructed surfaces while retaining fine details. For surface-based evaluation metrics, we use the marching cubes algorithm \cite{lorensen87} to extract meshes from volumetric representations.

\subsection{Recognition baselines}
\label{sec:recognition_baselines}

We implemented two straightforward baselines that approach the problem purely in terms of recognition. The first is based on clustering of the training shapes in conjunction with an image classifier; the second performs database retrieval.

\mypara{Clustering.}
In this baseline, we cluster the training shapes into $K$ sub-categories using the K-means algorithm~\cite{Macqueen1967}.
Since using $128^3$ voxelizations as feature vectors for clustering is too costly, we run the algorithm on $32^3$ voxelizations flattened into a vector.
Once the cluster assignments are determined, we switch back to working with high-resolution models.

Within each of the $K$ clusters, we calculate the mean shape as
\begin{equation}
    \hat{m}_k = \frac{1}{N_k}\sum_{n=0}^{N_k} v_n,
\end{equation}
where $v_n$ is one of the $N_k$ shapes belonging to the $k$-th cluster.
We threshold the mean shapes at $\tau_k$, where the optimal $\tau_k$ value is determined by maximizing the average IoU over the models belonging to the $k$-th cluster:
\begin{equation}
    \tau_k = \arg \max_{\tau} \frac{1}{N_k}\sum_{n=0}^{N_k}\textnormal{IoU}(\hat{m}_k > \tau, v_n),
\end{equation}
where the thresholding operation is applied per voxel. We enumerate 
$\tau$ in the interval $[0.05, 0.5]$ with a step size of $0.05$ to find the optimal threshold. We set $K=500$.

Since correspondences between images and 3D shapes are known for the training set, images can be readily matched with the respective cluster $k$.
Subsequently, we train a 1-of-$K$ classifier that assigns images to cluster labels.
At test time, we set the mean shape of the predicted cluster as the inferred solution.
For classification, we use the ResNet-50 architecture~\cite{he16}, pre-trained on the ImageNet dataset~\cite{deng09}, and fine-tuned for 30 epochs on our data.

\mypara{Retrieval.}
Our retrieval baseline is inspired by the work of Li~\etal~\cite{li15}, which learns to embed images and shapes in a joint space. The embedding space is constructed from the pairwise similarity matrix of all 3D shapes in the training set by compressing each row of the matrix to a low-dimensional descriptor via Multi-Dimensional Scaling~\cite{kruskal64} with Sammon mapping~\cite{sammon69}. To compute the similarity of two arbitrary shapes, Li \etal employ the lightfield descriptor~\cite{chen03}. To embed images in the space spanned by the shape descriptors, a convolutional network~\cite{krizhevsky12} is trained to map images to the descriptor given by the corresponding shape in the training set. During training, the network optimizes the Euclidean distance between predicted and ground-truth descriptors.

We adapt the work of Li~\etal in several ways. As with our clustering baseline, we determine the similarity between two shapes via the IoU of their $32^3$ voxel grid representation. We then compute a low-dimensional descriptor via principal component analysis.
We further use a larger descriptor (512 vs.\ 128) and a network with larger capacity (ResNet-50~\cite{he16}, pre-trained on ImageNet~\cite{deng09}, without fixing any layers during fine-tuning). 
Finally, instead of minimizing the Euclidean distance, we maximize the cosine similarity between descriptors during training.

\mypara{Oracle nearest neighbor.}
To gain more insight into the characteristics of the dataset, we evaluate an Oracle Nearest Neighbor (Oracle NN) baseline.
For each of the test 3D shapes, we find the closest shape from the training set in terms of IoU.
This method cannot be applied in practice, but gives an upper bound on how well a retrieval method can solve the task.

\begin{figure}[tbp]
\centering
\includegraphics[clip,trim=0 0 0 0, width=8.25cm]{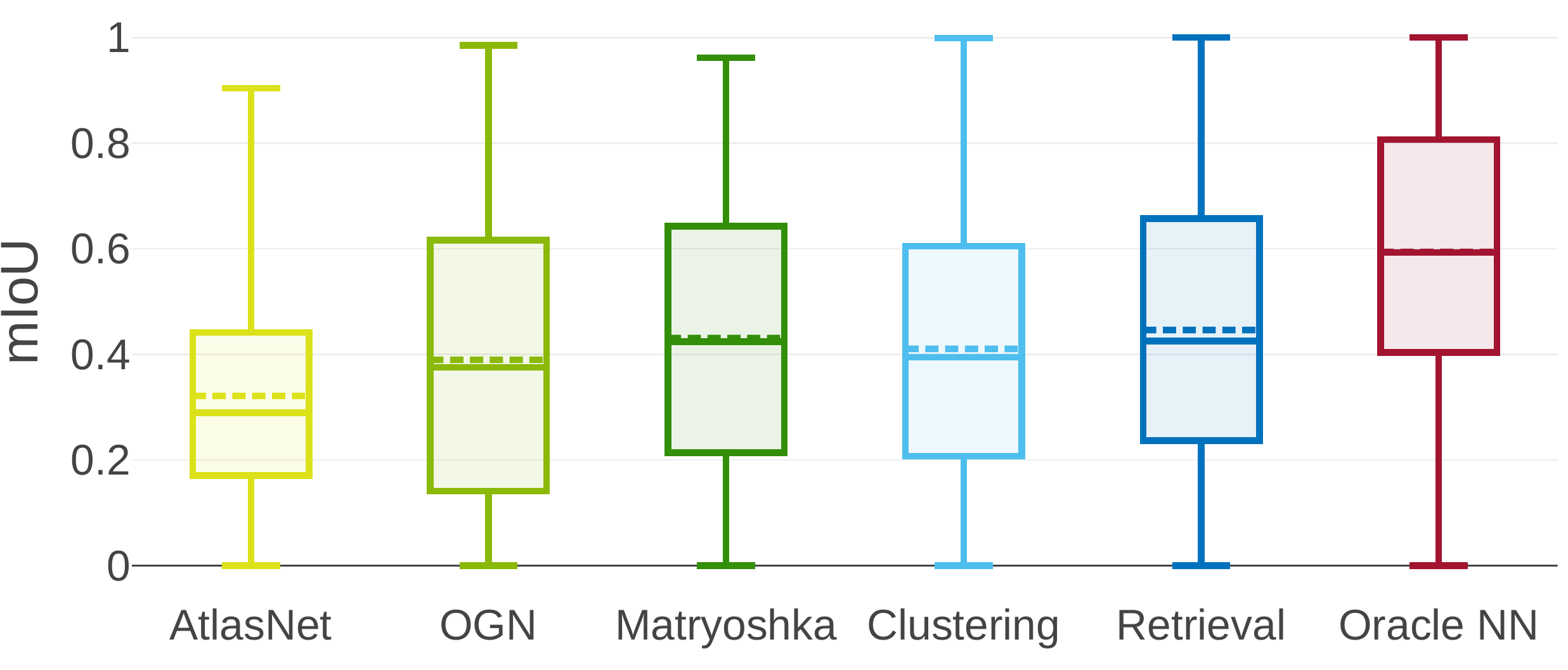}
\vspace{-1em}
\caption{Comparison by mean IoU over the dataset. The box corresponds to the second and third quartile. The solid line in the box depicts the median; the dashed line the mean. Whiskers mark the minimum and maximum values, respectively.}  
\label{fig:mean_iou}
\end{figure}

\subsection{Analysis}
We start by conducting a standard comparison of all methods in terms of their mean IoU scores. The results are summarized in Fig.~\ref{fig:mean_iou}.
We find that state-of-the-art methods, despite being backed by different architectures, perform at a remarkably similar level.
Interestingly, the retrieval baseline, a pure recognition method, outperforms all other approaches both in terms of mean and median IoU. The simple clustering baseline is competitive and outperforms both AtlasNet and OGN.
We further observe that a perfect retrieval method (Oracle NN) performs significantly better than all other methods.
Strikingly, the variance in the results is extremely high (between 35\% and 50\%) for all methods.
This implies that quantitative comparisons that rely solely on the mean IoU do not provide a full picture at this level of performance. To shed more light on the behavior of the methods, we proceed with a more detailed analysis.
\begin{figure*}
\centering
\includegraphics[clip,trim=0 3.5cm 0 0,width=17.5cm]{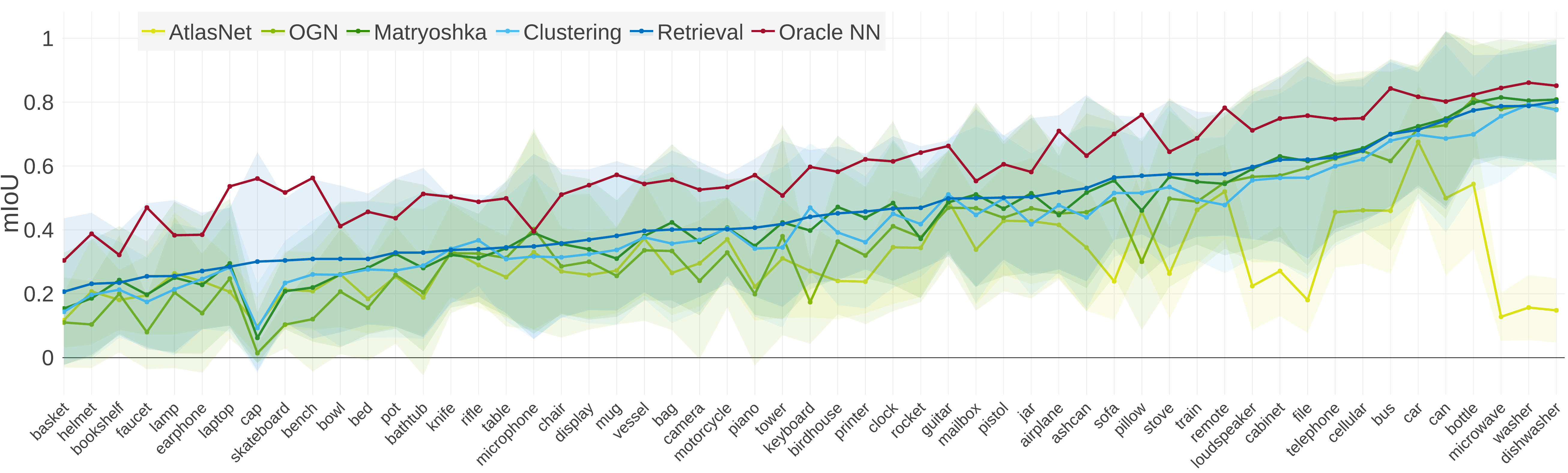}
\includegraphics[clip,trim=0 0 0 13.5cm,width=17.5cm]{figures/class_vs_performance/class_vs_performance.pdf}
\vspace{-2.0em}
\caption{Comparison by mIoU per class. Overall, the methods exhibit consistent relative performance across different classes.
The retrieval baseline produces the best reconstructions for the majority of classes.
The variance is high for all classes and methods.}
\label{fig:class_vs_performance}
\end{figure*}

\mypara{Per-class analysis.}
The similarity in average accuracy cannot be attributed to methods specializing in different subsets of classes.
In Fig.~\ref{fig:class_vs_performance} we observe consistent relative performance between methods across different classes.
The retrieval baseline achieves the best results for 30 out of 55 classes.
The classes are sorted from left to right in ascending order according to the performance of the retrieval baseline.
The variance is high for all classes and all \mbox{methods}.

\begin{figure}
\centering
\includegraphics[clip,trim=0 0 0 0,width=8.25cm]{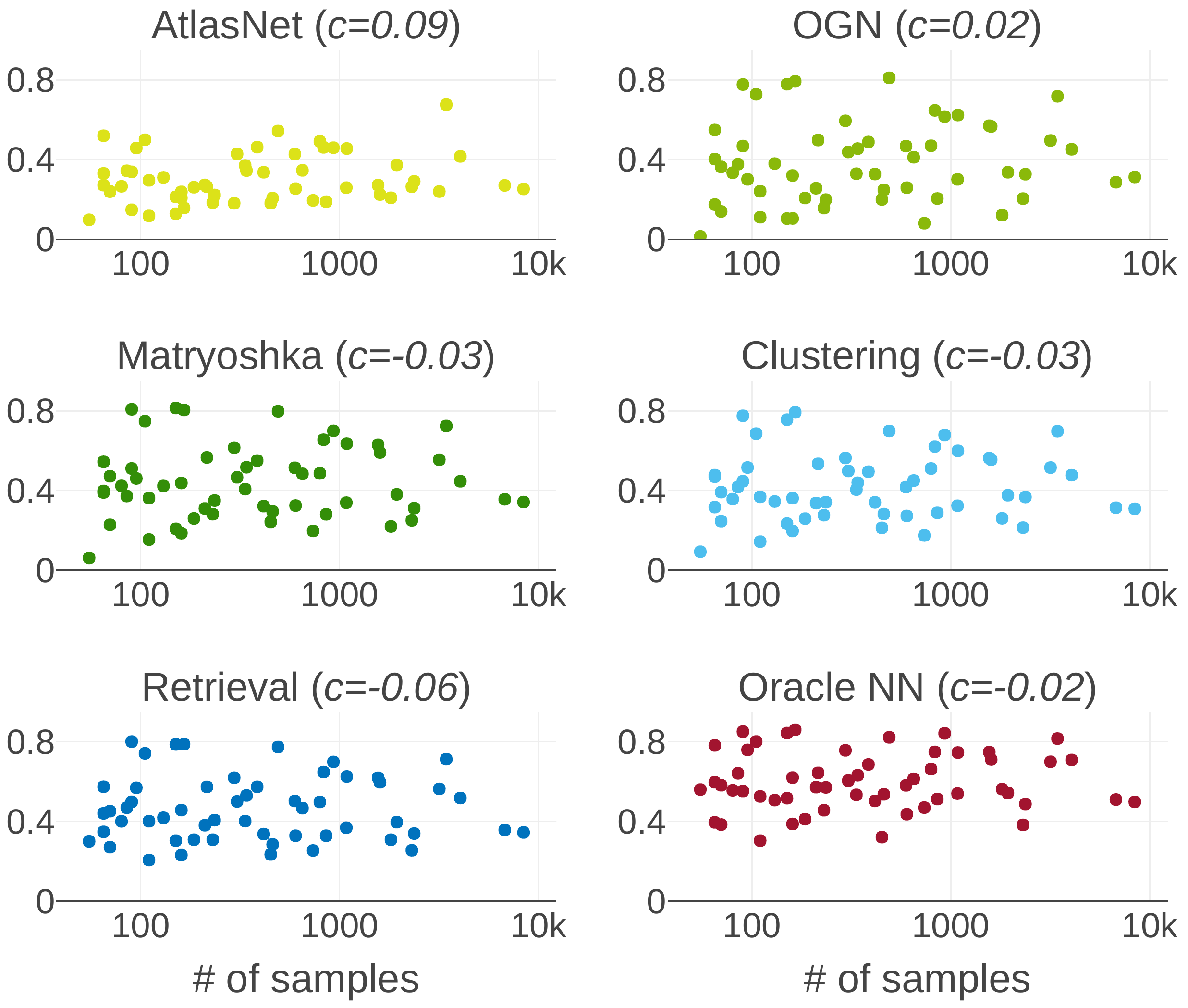}
\vspace{-0.8em}
\caption{mIoU versus number of training samples per class. We find no correlation between the number of samples within a class and the mIoU score for this class. The correlation coefficient $c$ is close to zero for all methods.
}
\label{fig:size_vs_performance}
\end{figure}

One might assume that the per-class performance depends on the number of training samples that are available for a class. 
However, we find no correlation  between the number of samples in a class and its mean IoU score; see Fig.~\ref{fig:size_vs_performance}.
The correlation coefficient between the two quantities is close to zero for all methods.
This implies that there is no justification for only using 13 out of the 55 classes, as was done in many \mbox{prior works ~\cite{choy16,fan17,groueix18,richter18,tatarchenko17,yan16}}.

\begin{figure*}
\begin{tabular}{@{}*{7}{c@{\hspace{0.2mm}}}c@{}}
\footnotesize Input & 
\footnotesize Ground truth & 
\footnotesize AtlasNet & 
\footnotesize OGN & 
\footnotesize Matryoshka & 
\footnotesize Clustering & 
\footnotesize Retrieval & 
\footnotesize Oracle NN\\ 
\includegraphics[width=0.125\linewidth]{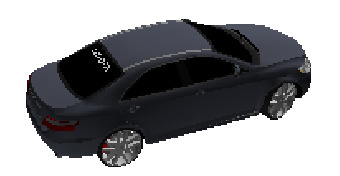} &
\includegraphics[clip,trim=0.3cm 2cm 1cm 2cm,width=0.12\linewidth]{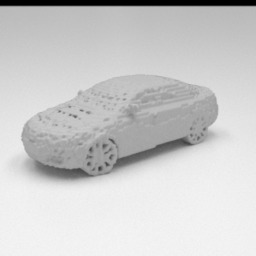} &
\begin{overpic}[clip,trim=0.3cm 2cm 1cm 2cm,width=0.12\linewidth]{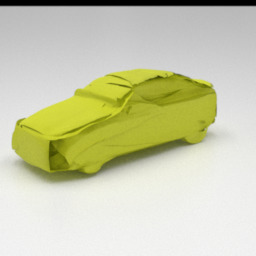}
 \put (75,7) {\scriptsize 0.69}
\end{overpic}&
\begin{overpic}[clip,trim=0.3cm 2cm 1cm 2cm,width=0.12\linewidth]{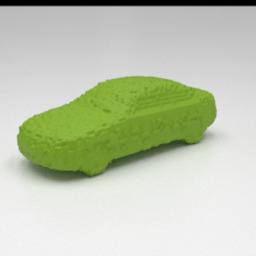}
 \put (75,7) {\scriptsize 0.78}
\end{overpic} &
\begin{overpic}[clip,trim=0.3cm 2cm 1cm 2cm,width=0.12\linewidth]{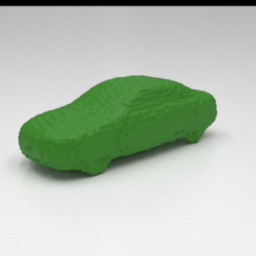}
 \put (75,7) {\scriptsize 0.77}
\end{overpic} &
\begin{overpic}[clip,trim=0.3cm 2cm 1cm 2cm,width=0.12\linewidth]{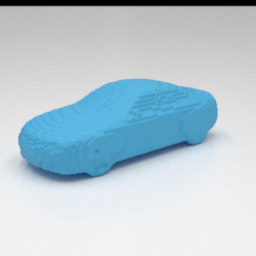}
 \put (75,7) {\scriptsize 0.73}
\end{overpic} &
\begin{overpic}[clip,trim=0.3cm 2cm 1cm 2cm,width=0.12\linewidth]{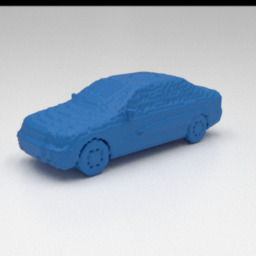}
 \put (75,7) {\scriptsize 0.75}
\end{overpic} &
\begin{overpic}[clip,trim=0.3cm 2cm 1cm 2cm,width=0.12\linewidth]{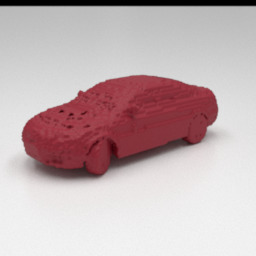}
 \put (75,7) {\scriptsize 0.93}
\end{overpic}\\[-2.5pt]
\includegraphics[width=0.125\linewidth]{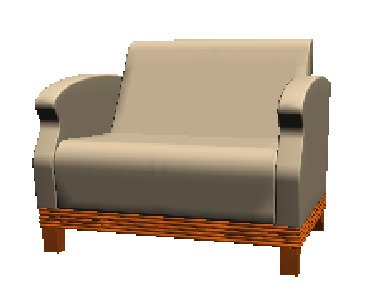} &
\includegraphics[clip,trim=0 1cm 0 0,width=0.12\linewidth]{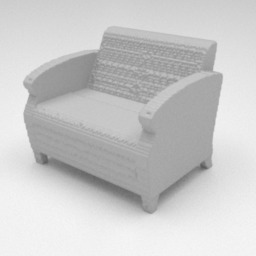} &
\begin{overpic}[clip,trim=0 1cm 0 0,width=0.12\linewidth]{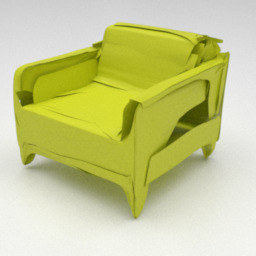}
 \put (75,7) {\scriptsize 0.15}
\end{overpic}&
\begin{overpic}[clip,trim=0 1cm 0 0,width=0.12\linewidth]{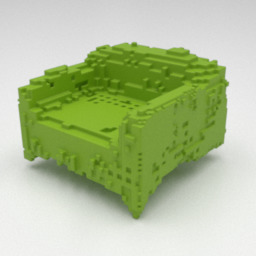}
 \put (75,7) {\scriptsize 0.59}
\end{overpic} &
\begin{overpic}[clip,trim=0 1cm 0 0,width=0.12\linewidth]{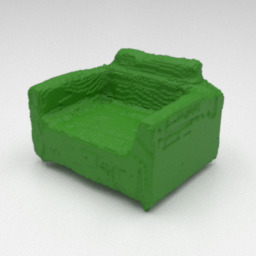}
 \put (75,7) {\scriptsize 0.71}
\end{overpic} &
\begin{overpic}[clip,trim=0 1cm 0 0,width=0.12\linewidth]{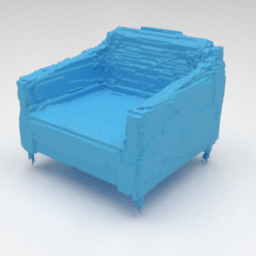}
 \put (75,7) {\scriptsize 0.58}
\end{overpic} &
\begin{overpic}[clip,trim=0 1cm 0 0,width=0.12\linewidth]{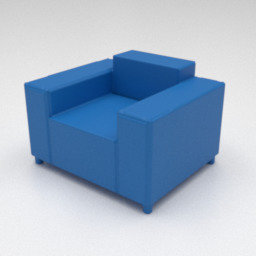}
 \put (75,7) {\scriptsize 0.68}
\end{overpic} &
\begin{overpic}[clip,trim=0 1cm 0 0,width=0.12\linewidth]{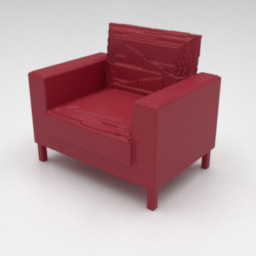}
 \put (75,7) {\scriptsize 0.72}
\end{overpic}\\[-2.5pt]
\includegraphics[width=0.125\linewidth]{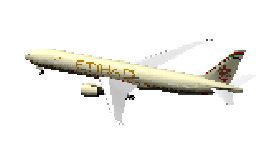} &
\includegraphics[clip,trim=0.8cm 2cm 1cm 2cm,width=0.12\linewidth]{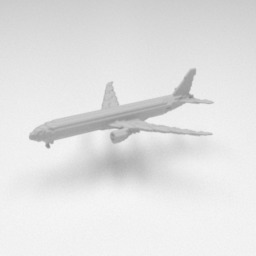} &
\begin{overpic}[clip,trim=0.8cm 2cm 1cm 2cm,width=0.12\linewidth]{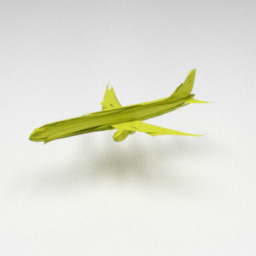}
 \put (75,7) {\scriptsize 0.62}
\end{overpic}&
\begin{overpic}[clip,trim=0.8cm 2cm 1cm 2cm,width=0.12\linewidth]{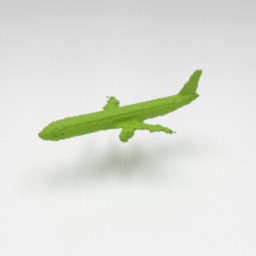}
 \put (75,7) {\scriptsize 0.77}
\end{overpic} &
\begin{overpic}[clip,trim=0.8cm 2cm 1cm 2cm,width=0.12\linewidth]{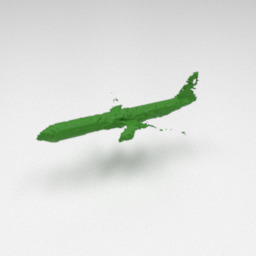}
 \put (75,7) {\scriptsize 0.67}
\end{overpic} &
\begin{overpic}[clip,trim=0.8cm 2cm 1cm 2cm,width=0.12\linewidth]{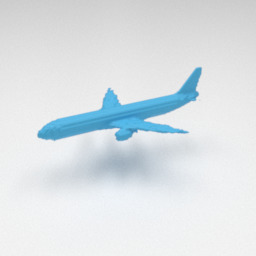}
 \put (75,7) {\scriptsize 0.81}
\end{overpic} &
\begin{overpic}[clip,trim=0.8cm 2cm 1cm 2cm,width=0.12\linewidth]{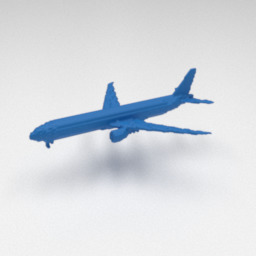}
 \put (75,7) {\scriptsize 0.92}
\end{overpic} &
\begin{overpic}[clip,trim=0.8cm 2cm 1cm 2cm,width=0.12\linewidth]{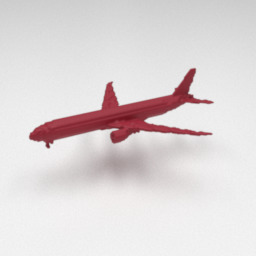}
 \put (75,7) {\scriptsize 0.98}
\end{overpic}\\[-2.5pt]
\includegraphics[width=0.10\linewidth]{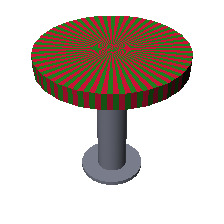} &
\includegraphics[clip,trim=0 0.8cm 0 0.5cm,width=0.12\linewidth]{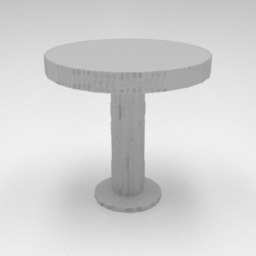} &
\begin{overpic}[clip,trim=0 0.8cm 0 0.5cm,width=0.12\linewidth]{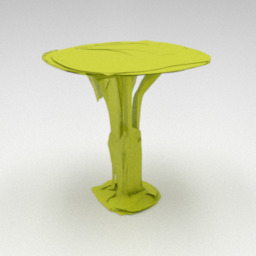}
 \put (75,7) {\scriptsize 0.26}
\end{overpic}&
\begin{overpic}[clip,trim=0 0.8cm 0 0.5cm,width=0.12\linewidth]{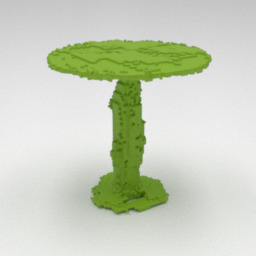}
 \put (75,7) {\scriptsize 0.42}
\end{overpic} &
\begin{overpic}[clip,trim=0 0.8cm 0 0.5cm,width=0.12\linewidth]{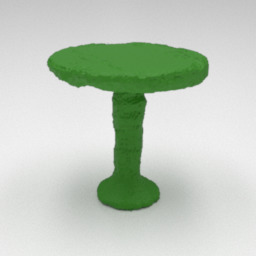}
 \put (75,7) {\scriptsize 0.69}
\end{overpic} &
\begin{overpic}[clip,trim=0 0.8cm 0 0.5cm,width=0.12\linewidth]{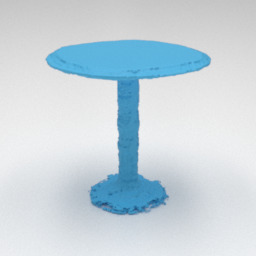}
 \put (75,7) {\scriptsize 0.44}
\end{overpic} &
\begin{overpic}[clip,trim=0 0.8cm 0 0.5cm,width=0.12\linewidth]{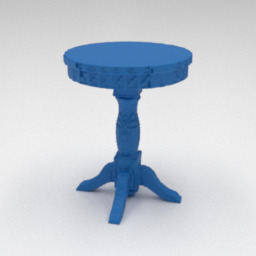}
 \put (75,7) {\scriptsize 0.39}
\end{overpic} &
\begin{overpic}[clip,trim=0 0.8cm 0 0.5cm,width=0.12\linewidth]{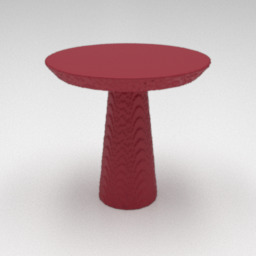}
 \put (75,7) {\scriptsize 0.47}
\end{overpic}
\end{tabular}
\vspace{-1em}
\caption{Qualitative results. Our clustering baseline produces shapes at a quality comparable to state-of-the-art approaches. Our retrieval baseline returns high-fidelity shapes by design, although details may not be correct.
Numbers in the bottom right corner of each sample indicate the IoU.}
\label{fig:qualitative}
\end{figure*}

The quantitative results are backed by qualitative results shown in Fig.~\ref{fig:qualitative}.
For most classes, there is no significant visual difference between the predictions of the decoder-based methods and our clustering baseline.
Clustering fails when the sample is far from the mean shape of the cluster, or when the cluster itself cannot be described well by the mean shape (this is often the case for chairs or tables because of thin structures that get averaged out in the mean shape).
The predictions of the retrieval baseline look more appealing in most cases due to the presence of fine details, even though these details are not necessarily correct.
We provide additional qualitative results in the supplementary material.

\begin{figure*}
\vspace{0.5em}
\begin{tabular}{c c}
\centering
\includegraphics[width=13cm, trim=0.25cm 0 0.4cm 0,clip]{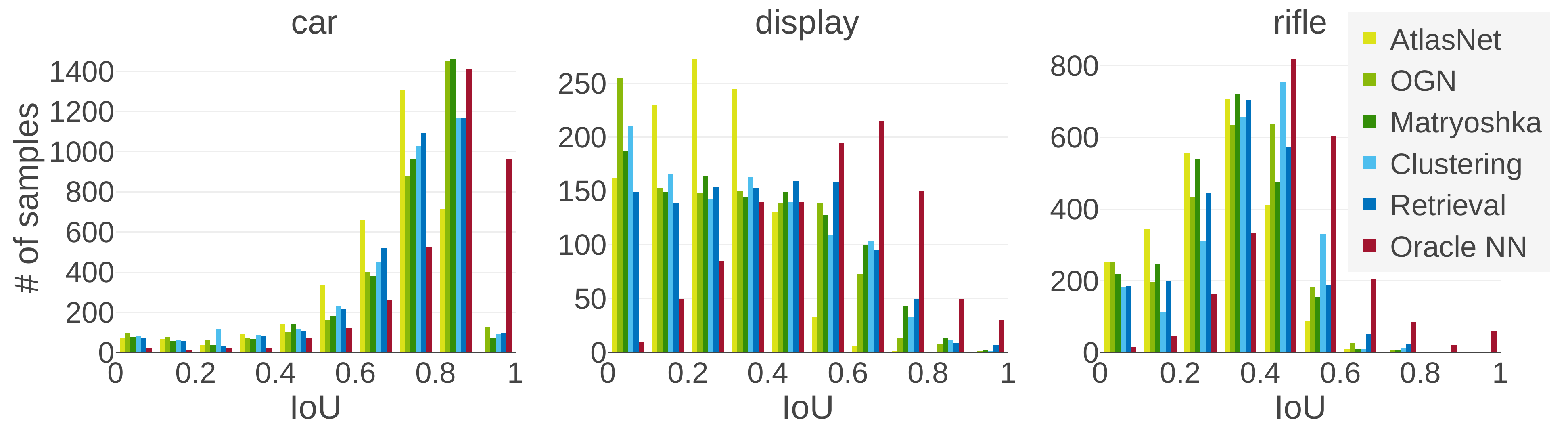} & \hspace{-8pt}
\includegraphics[width=4.0cm, trim=0.31cm 0 0 0,clip]{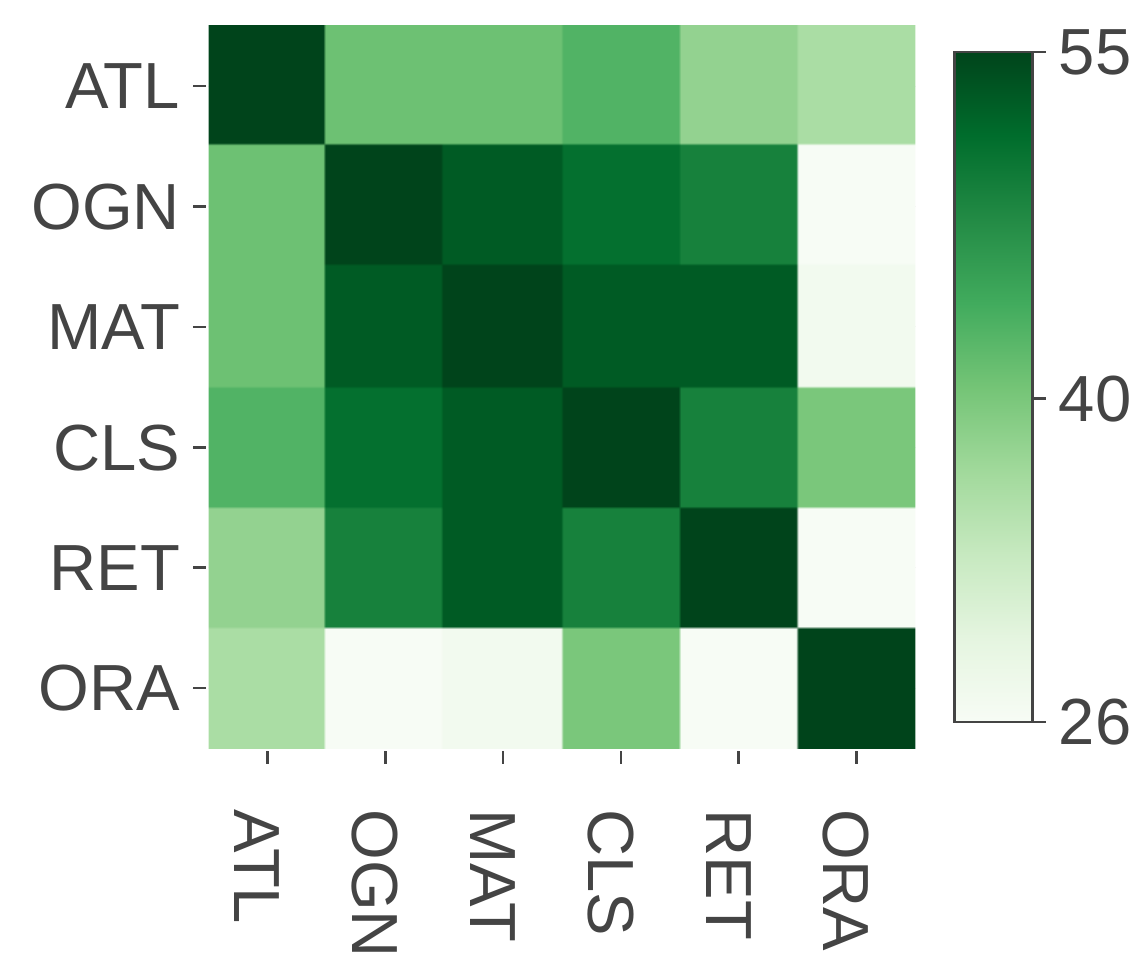}
\end{tabular}
\vspace{-1.2em}
\caption{Left: Distribution of IoUs for selected classes. Within-class distributions for decoder-based methods and explicit recognition baselines are similar. The distributions of the Oracle NN differ for most of the classes. Right: A heat map of the number of classes for which the pairwise Kolmogorov-Smirnov test fails to reject the null hypotheses of the two distributions being the same.}
\label{fig:histograms}
\end{figure*}

\mypara{Statistical evaluation.}
To further investigate the hypothesis that convolutional networks bypass true reconstruction via image recognition, we visualize the histograms of IoU scores for individual object classes in Fig.~\ref{fig:histograms}.
For histograms of all 55 classes we refer to the supplementary material.
Although the distributions differ between classes, the within-class distributions of decoder-based methods and recognition baselines are surprisingly similar.

For reference, we also plot the results of the Oracle NN baseline, which, for many classes, differs substantially.
To verify this observation rigorously, we perform the Kolmogorov-Smirnov test \cite{kolmogorov51} on the 50-binned versions of the histograms for all classes and all pairs of methods.
The null hypothesis assumes that two distributions exhibit no statistically significant difference.
We visualize the results of the test in the rightmost part of Fig.~\ref{fig:histograms}.
Every cell of the heat map shows the number of classes for which the statistical test does not allow to reject the null hypothesis, \ie, where the p-value is larger than $0.05$.
We find that for decoder-based methods and recognition baselines the null hypothesis cannot be rejected for the vast majority of classes.

\section{Problems}
\label{sec:problems}
In the preceding section we provided evidence that current methods for single-view 3D object reconstruction predominantly rely on recognition.
Here we discuss aspects of popular experimental procedures that may need to be reconsidered to elicit more detailed reconstruction behavior from the models.

\subsection{Choice of coordinate system}

The vast majority of existing methods predict output shapes in an object-centered coordinate system, which aligns objects of the same semantic category to a common orientation. Aligning objects this way makes it particularly easy to find spatial regularities. It encourages learning-based approaches to recognize the object category first, and refine the shape later if at all.

Shin \etal~\cite{shin18} studied how the choice of coordinate frames affects reconstruction performance and generalization abilities of learning-based methods, comparing object-centered and viewer-centered coordinate frames. They found that a viewer-centered frame leads to significantly better generalization to object classes that were not seen during training, a result that can only be achieved when a method operates in a geometric reconstruction regime.

\begin{figure}[htbp]
\centering
\includegraphics[clip,trim=0 0 0 0, width=8.25cm]{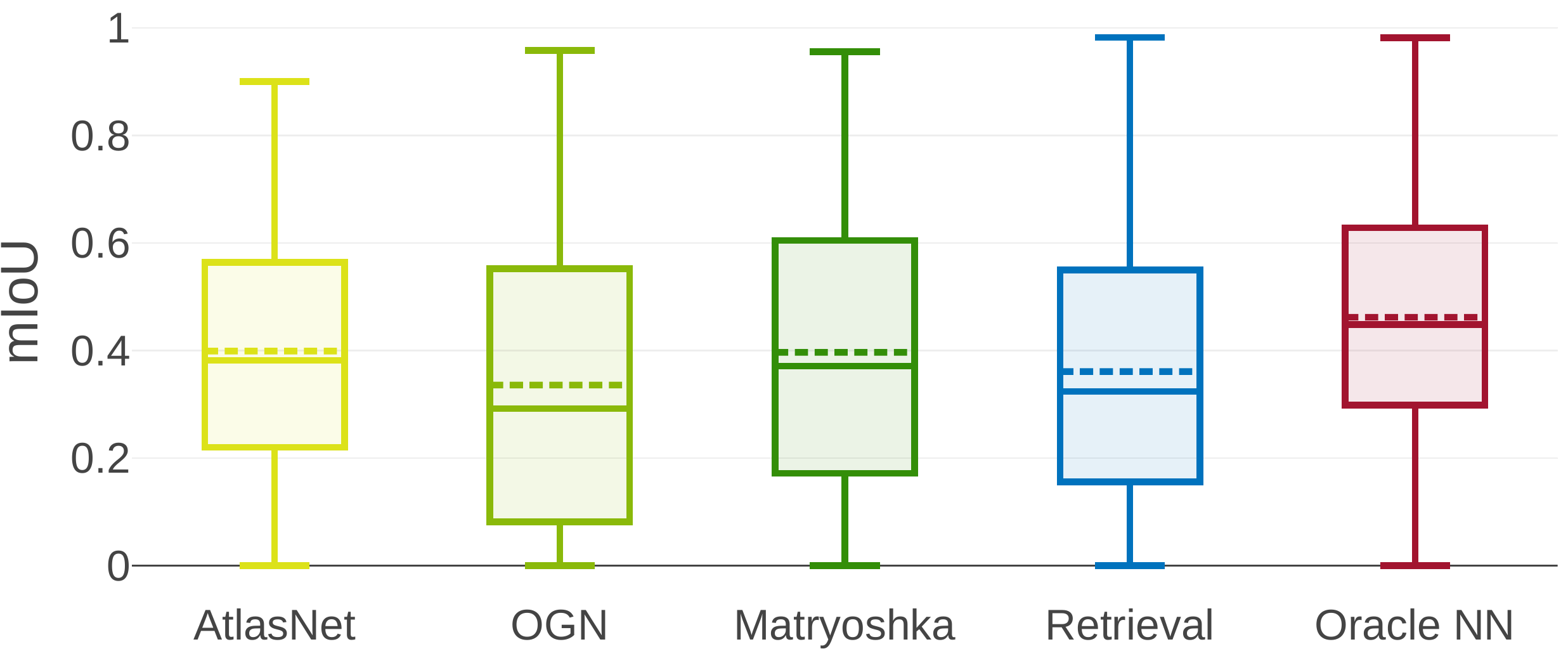}
\vspace{-1em}
\caption{Mean IoU in viewer-centered mode. The retrieval baseline does not perform as well in this mode.}
\label{fig:mean_iou_viewer_centered}
\end{figure}

To validate these conclusions, we repeated the experimental evaluation (Sec.~\ref{sec:conventional_evaluation}) in a viewer-centered coordinate frame.
We attempted to extend the clustering baseline with a viewpoint prediction network which would regress the azimuth and elevation angles of the camera w.r.t. the canonical frame.
This naive approach failed because the canonical frame has a different meaning for each object class, implying that the viewpoint network needs to use class information in order to solve the task.
For the retrieval baseline, we retrained the method, treating each training view as a separate sample.
To avoid artifacts from rotating voxelized shapes, we synthesized ground-truth shapes by rotating and then voxelizing the original meshes, resulting in a distinct target shape for each view of each object. Results are shown in Fig.~\ref{fig:mean_iou_viewer_centered}, where we observe a mild decrease in performance for OGN and Matryoshka networks, and a larger drop for the retrieval baseline.
For the retrieval setting, the viewer-centered setup is computationally more demanding, as different views of the same object now refer to different shapes to be retrieved. Consequently, less learning capacity is available for each individual object.

\subsection{Evaluation metric}

\mypara{Intersection over union.}
The mean IoU is commonly used as the primary quantitative measure for benchmarking single-view reconstruction approaches.
This can be problematic if it is used as the sole metric to argue for the merits of an approach, since it is only indicative of the quality of a predicted shape if it reaches sufficiently high values. Low to mid-range scores indicate a significant discrepancy between two shapes.
\begin{figure*}
\begin{tabular}{@{}*{5}{c@{\hspace{0.2mm}}}c@{}}
\footnotesize Source &
\footnotesize 0.20 &
\footnotesize 0.40 &
\footnotesize 0.59 &
\footnotesize 0.81 &
\footnotesize 0.91\\
\includegraphics[clip,trim=0 1cm 0 1.5cm, width=0.165\linewidth]{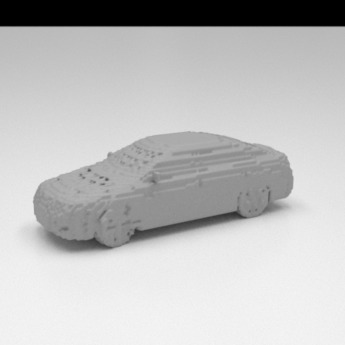} &
\includegraphics[clip,trim=0 1cm 0 1.5cm, width=0.165\linewidth]{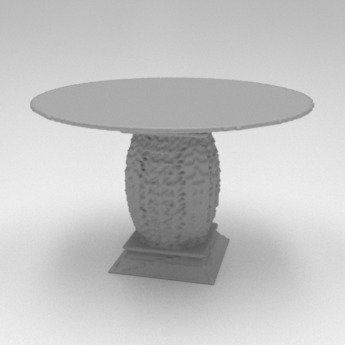} &
\includegraphics[clip,trim=0 1.5cm 0 1cm, width=0.165\linewidth]{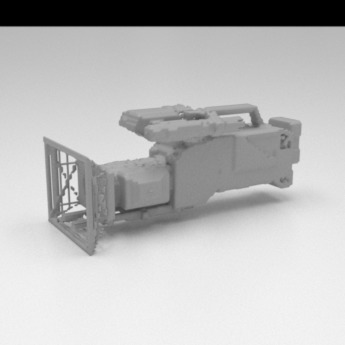} &
\includegraphics[clip,trim=0 1cm 0 1.5cm, width=0.165\linewidth]{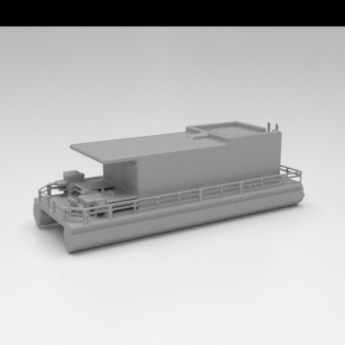} &
\includegraphics[clip,trim=0 1cm 0 1.5cm, width=0.165\linewidth]{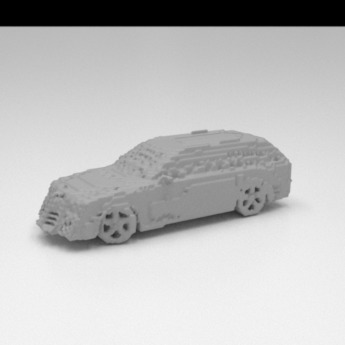} &
\includegraphics[clip,trim=0 1cm 0 1.5cm, width=0.165\linewidth]{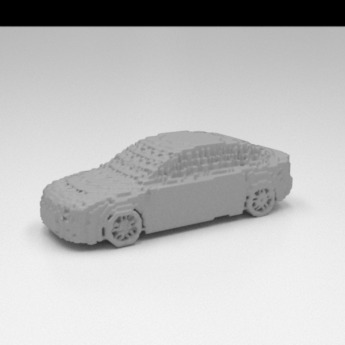}
\end{tabular}
\vspace{0.5em}
\vspace{-1.2em}
\caption{IoU between a source shape and various target shapes. Low to mid-range IoU values are a poor indicator of shape similarity.}
\label{fig:iou_example}
\end{figure*}

An example is shown in Fig.~\ref{fig:iou_example}, which compares a car model to different shapes in the dataset and illustrates their similarity in terms of IoU scores. As shown in the figure, even an IoU of 0.59 allows for considerable deviation from the ground-truth shape. For reference, note that 75\% of the predictions by the best performing approach, our retrieval baseline, have an IoU below 0.66; 50\% are below 0.43 \mbox{(\cf Fig. \ref{fig:mean_iou})}.

All information about an object's shape is situated on its surface.
However, for voxel-based representations with a solid interior, the IoU is dominated by the interior parts of objects.
As a consequence, even seemingly high IoU values may poorly reflect the actual surface similarity.

Moreover, while IoU can easily be evaluated for a volumetric representation, there is no straightforward way to evaluate it for point clouds.
A good measure should allow comparing different 3D representations within the same unified framework.
Point-based measures are most suitable for this, because a point cloud can be obtained from any other 3D representation via (a)~surface point sampling for meshes, (b)~per-pixel reprojection for depth maps, or (c)~running the marching cubes algorithm followed by point sampling for voxel grids.

\begin{figure}
\begin{tabular}{@{}*{2}{c@{\hspace{0.2mm}}}c@{}}
\footnotesize Source &
\footnotesize CD = 0.21 &
\footnotesize CD = 0.15 \\
\includegraphics[clip,trim=0 1cm 0 1.5cm, width=0.33\linewidth]{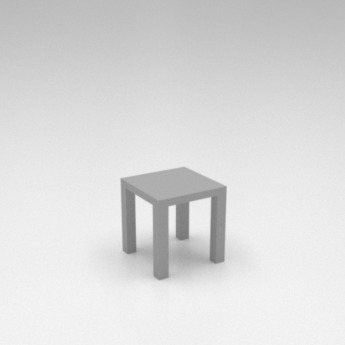} &
\includegraphics[clip,trim=0 1cm 0 1.5cm, width=0.33\linewidth]{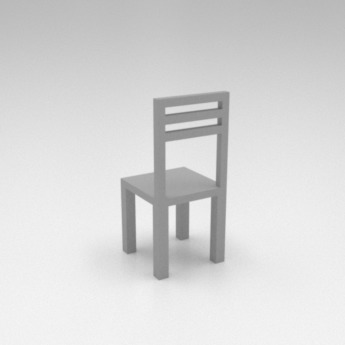} &
\includegraphics[clip,trim=0 1cm 0 1.5cm, width=0.33\linewidth]{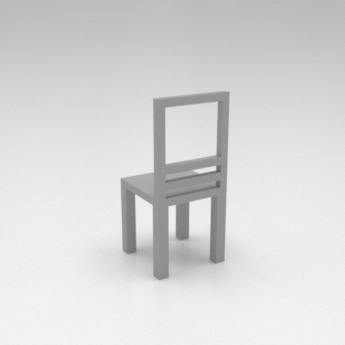}
\end{tabular}
\vspace{0.5em}
\vspace{-1em}
\caption{The Chamfer distance is sensitive to outliers. Compared to the source, both target shapes exhibit non-matching parts that are equally wrong. While the F@1\% is $0.56$ for both shapes, the Chamfer distance differs significantly.}
\label{fig:cd_example}
\end{figure}

\mypara{Chamfer distance.}
Some recent methods use the Chamfer Distance (CD) for evaluation \cite{fan17,groueix18,sun18}.
Although it is defined on point clouds and by design satisfies the requirement of being applicable (after conversion) to different 3D representations, it is a problematic measure because of its sensitivity to outliers.
Consider the example in Fig.~\ref{fig:cd_example}.
Both target chairs perfectly match the source chair in the lower part and are completely wrong in the upper part.
However, according to the CD score, the second target is much better than the first.
As this example shows, the CD measure can be significantly perturbed by the geometric layout of outliers. It is affected by how far the outliers are from the reference shape.
We argue that in order to reliably reflect real reconstruction performance, a good quantitative measure should be robust to the detailed geometry of outliers.

\mypara{F-score.}
Motivated by the insight that both IoU and CD can be misleading, we propose to use the F-score \cite{knapitsch17}, an established and easily interpretable metric that is actively used in the multi-view 3D reconstruction community.
The F-score explicitly evaluates the distance between object surfaces and is defined as the
harmonic mean between precision and recall. Precision measures the accuracy of the reconstruction by counting the percentage of reconstructed points that lie within a certain distance to the ground truth. Recall measures the completeness of the reconstruction by counting the percentage of points on the ground truth that lie within a certain distance to the reconstruction. 
The strictness of the F-score can be controlled by varying the distance threshold~$d$.
The metric has an intuitive interpretation: the percentage of points (or surface area) that was reconstructed correctly.

\begin{figure*}
\hspace*{-2mm}\begin{tabular}{cc}
\includegraphics[width=0.49\linewidth]{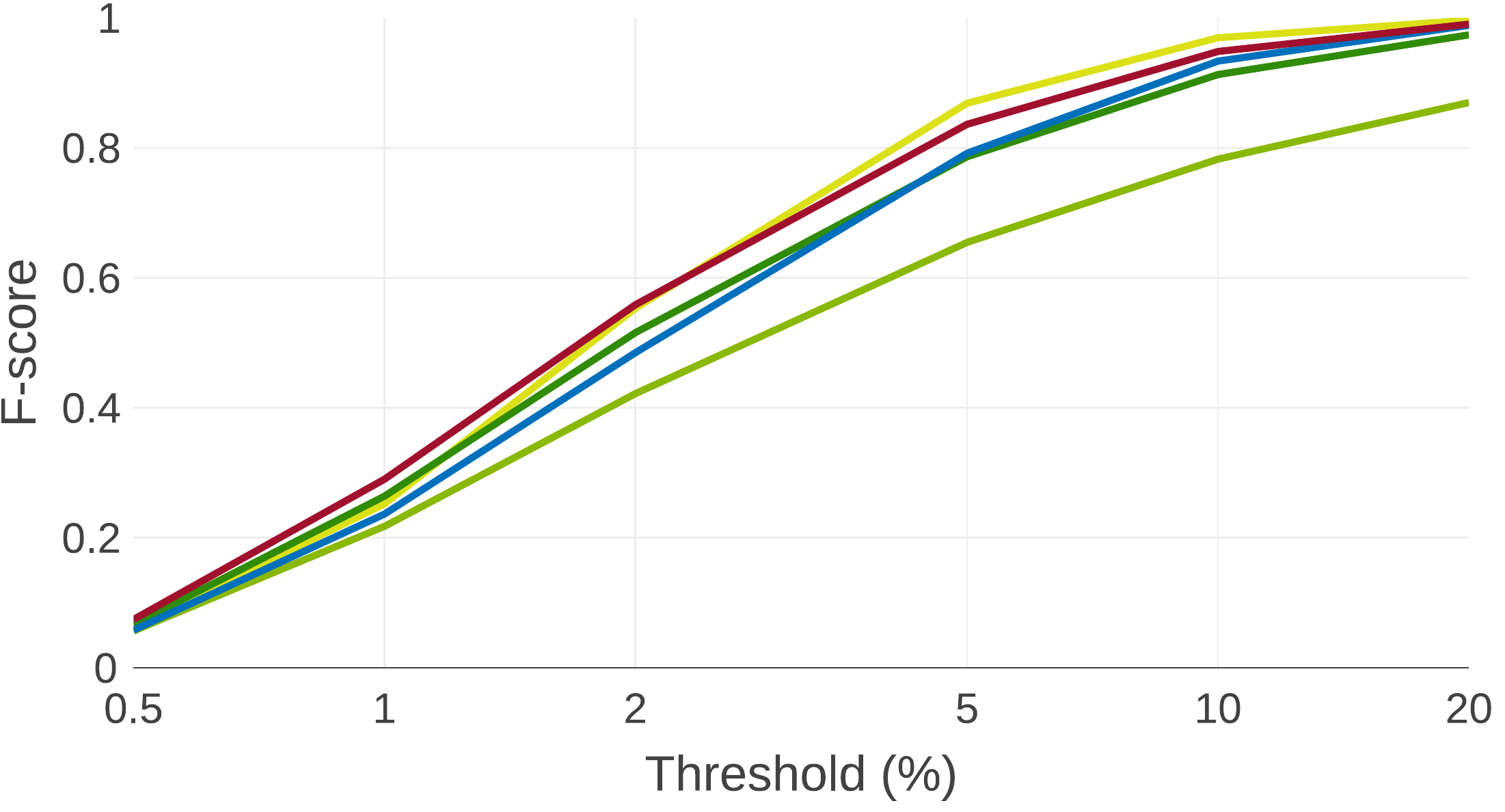} &
\includegraphics[width=0.49\linewidth]{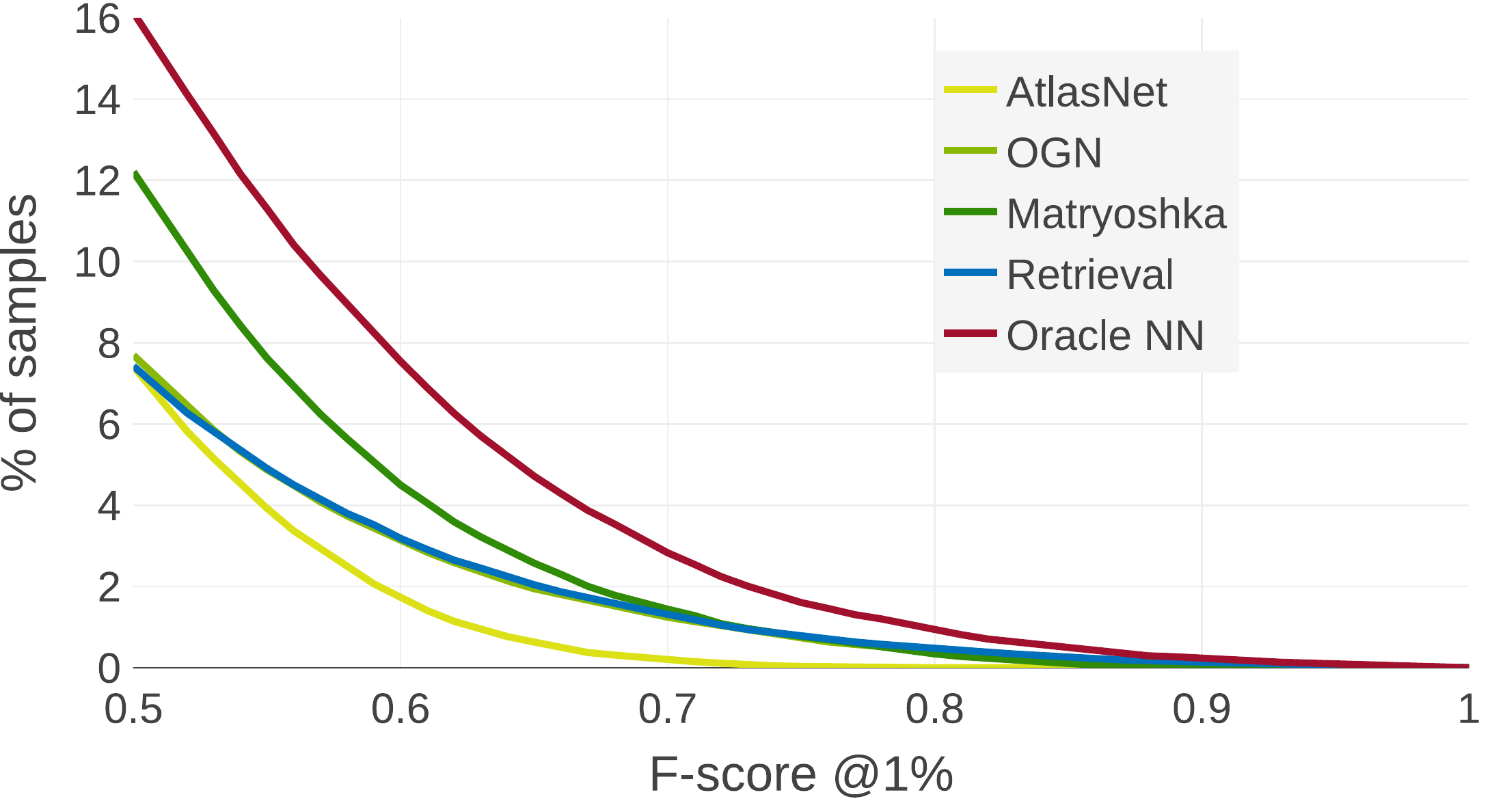}
\end{tabular}
\vspace{-1em}
\caption{F-score statistics in viewer-centered mode. Left: F-score for varying distance thresholds. Right: percentage of reconstructions with F-score above a value specified on the horizontal axis, with a distance threshold $d = 1\%$.}
\label{fig:rocs}
\end{figure*}

We plot the F-score of viewer-centered reconstructions for different distance thresholds $d$ in Fig.~\ref{fig:rocs} (left).
At $d=2\%$ of the side length of the reconstructed volume, the absolute F-score values are in the same range as the current mIoU scores, which, as we argued before, is not indicative of the prediction quality.
We therefore suggest evaluating the F-score at distance thresholds of $1\%$ and below.

In Fig.~\ref{fig:rocs} (right), we show the percentage of models with an F-score of 0.5 or higher at a threshold ${d=1\%}$.
Only a small number of shapes is reconstructed accurately, indicating that the task is still far from solved.
Our retrieval baseline is no longer a clear winner, further showing that a reasonable solution in viewer-centered mode is harder to get using a pure recognition method.

\begin{figure}
\centering
\begin{tabular}{c c}
\includegraphics[clip, trim=0 0 0 0,width=0.46\linewidth]{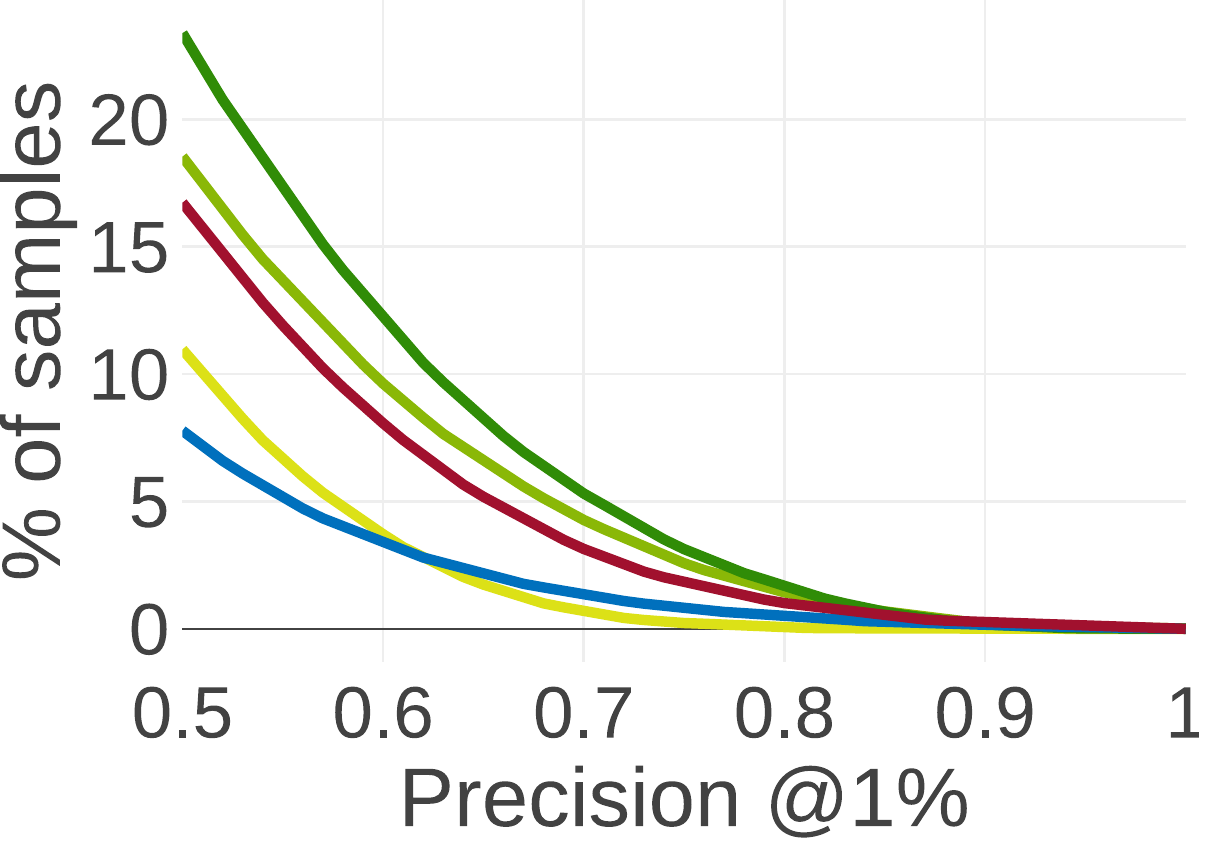} &
\includegraphics[clip, trim=0 0 0 0,width=0.46\linewidth]{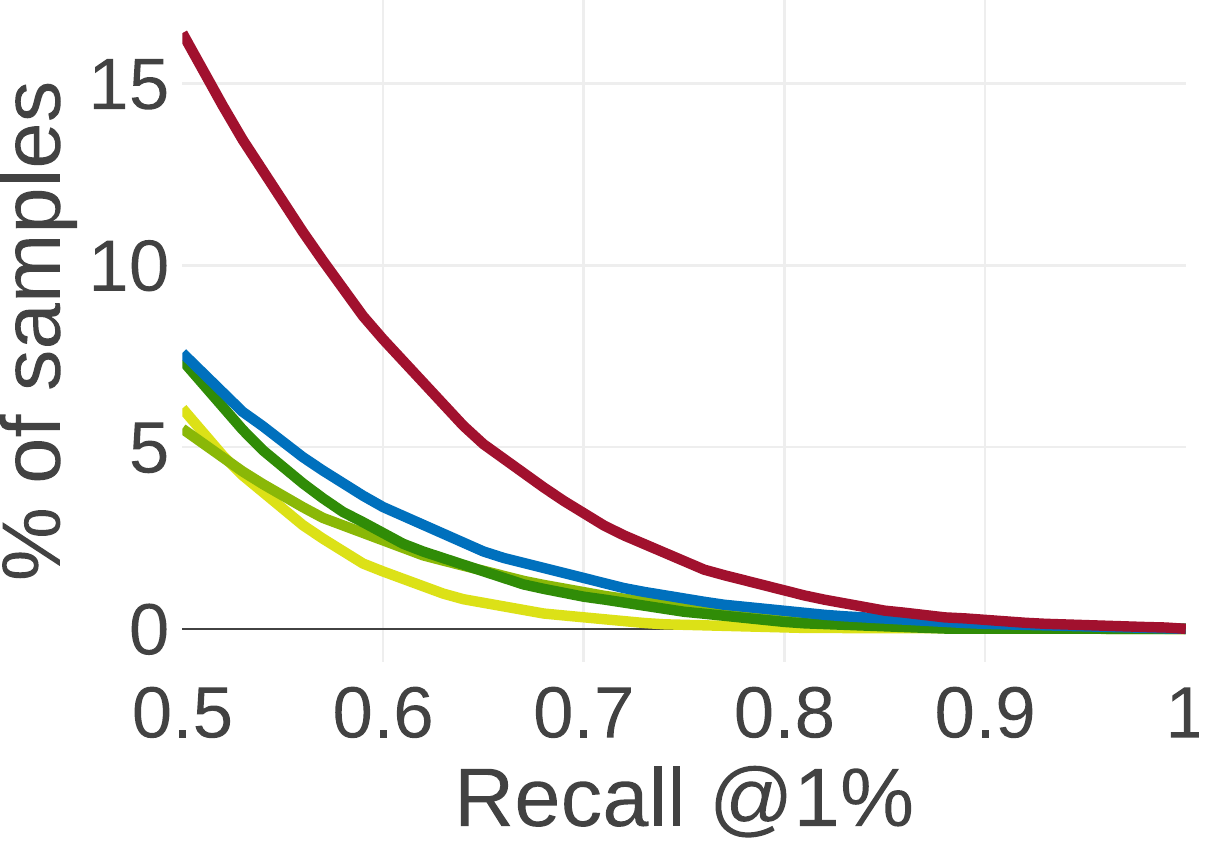}
\end{tabular}
\vspace{-1em}
\caption{Percentage of samples with precision (left) and recall (right) of 0.5 or higher. Existing CNN-based methods show good precision but miss parts of objects, which results in lower recall.}
\label{fig:precision_recall}
\end{figure}

We observe that AtlasNet often produces qualitatively good surfaces. It even outperforms the Oracle NN baseline on more liberal (above $2\%$) thresholds, as shown in Fig.~\ref{fig:rocs} (left). Perceptually,  humans tend to judge quality by global and semi-global features and tolerate if parts are slightly wrong in position or shape.
We observe that AtlasNet, which was trained to optimize surface correspondence, rarely completely misses parts of the model, but tends to produce poorly localized parts. This is reflected in the high-performance range analysis, shown in Fig.~\ref{fig:rocs} (right), where AtlasNet trails all other approaches.

Analyzing precision and recall separately provides additional insights into each method's behavior.
In Fig.~\ref{fig:precision_recall} we see that OGN and Matryoshka Networks outperform Oracle NN in terms of precision. However, both Oracle NN and the retrieval baseline show higher recall.
This is supported by qualitative observations that OGN and Matryoshka Networks tend to produce incomplete models.

Both recall and precision can be easily visualized to gain further insights, as illustrated in Fig.~\ref{fig:precision_recall_qual}.

\begin{figure}
\vspace{3mm}
\begin{tabular}{c c}
\begin{overpic}[width=0.46\linewidth]{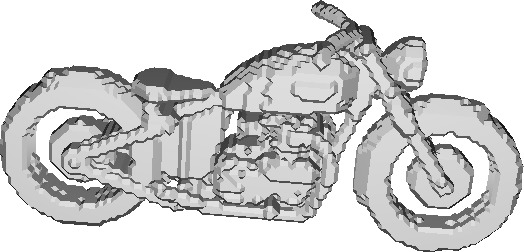}
    \put (42,50) {\footnotesize Retrieval}
\end{overpic} &
\begin{overpic}[width=0.46\linewidth]{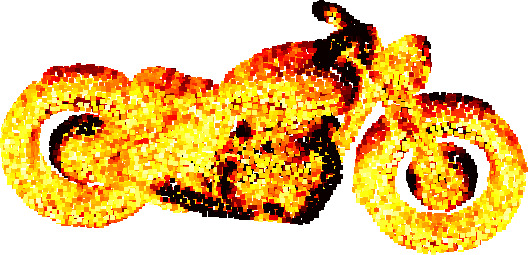}
    \put (42,50) {\footnotesize Precision}
\end{overpic}\vspace{5pt}\\
\begin{overpic}[width=0.46\linewidth]{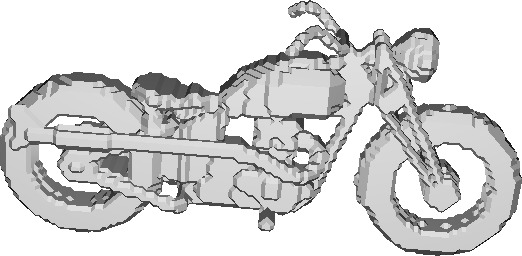}
    \put (36,50) {\footnotesize Ground truth}
\end{overpic} &
\begin{overpic}[width=0.46\linewidth]{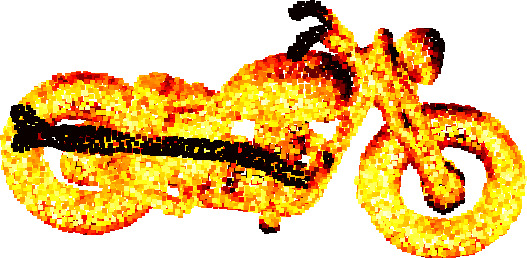}
    \put (45,50) {\footnotesize Recall}
\end{overpic}\\
& \begin{overpic}[width=0.43\linewidth]{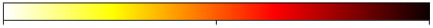}
    \put (0,-4) {\tiny 0\%}
    \put (49,-4) {\tiny 1\%}
    \put (98,-4) {\tiny 2\%}
\end{overpic} \hspace{5pt}\\
\end{tabular}
\vspace{0em}
\caption{Visualizing precision and recall provides detailed information about which object parts were reconstructed correctly. Colors encode the normalized distance between shapes (as used for the distance threshold).}
\label{fig:precision_recall_qual}
\end{figure}

\subsection{Dataset}
The problem of networks finding a semantic shortcut solution is closely related to the choice of training data.
The ShapeNet dataset has been used extensively because of its size.
However, its particular composition -- single objects of representative types, aligned to a canonical reference frame -- enables recognition models to masquerade as reconstruction.
In Fig.~\ref{fig:mean_iou}, we demonstrate that a retrieval solution (Oracle NN) outperforms all other methods on this dataset, \ie, the test data can be explained by simply retrieving models from the training set. This indicates a critical problem in using ShapeNet to evaluate 3D reconstruction: for a typical shape in the test set, there is a very similar shape in the training set. In effect, the train/test split is contaminated, because so many shapes within a class are similar. A reconstruction model evaluated on ShapeNet does not need to actually perform reconstruction: it merely needs to retrieve a similar shape from the training set.

\section{Conclusion}
\label{sec:conclusion}
In this paper, we reasoned about the spectrum of approaches to single-view 3D reconstruction, spanned by reconstruction and recognition. We introduced two baselines, classification and retrieval, which leverage only recognition. We showed that the simple retrieval baseline outperforms recent state-of-the-art methods. Our analysis indicates that state-of-the-art approaches to single-view 3D reconstruction primarily perform recognition rather than reconstruction. We identify aspects of common experimental procedures that elicit this behavior and make a number of recommendations, including the use of a viewer-centered coordinate frame and a robust and informative evaluation measure (the F-score). Another critical problem, the dataset composition, is identified but left unaddressed. We are working towards remedying this in a subsequent work.

\section*{Acknowledgements}

We thank Jaesik Park for his help with F-score evaluation. We also thank Max Argus and Estibaliz G\'{o}mez for valuable discussions and suggestions. This project used the Open3D library~\cite{Zhou2018}.

\clearpage

\balance

{\small
\bibliographystyle{ieee}
\bibliography{paper}
}

\clearpage

\section*{Appendix}

\appendix
\label{sec:supplementary}
\section{Metrics and evaluation protocol}
For completeness, we provide the definitions of the evaluation metrics used and additional details for converting different shape representations for evaluation.

\subsection{Intersection over union (IoU)}

In the context of 3D shape reconstruction, the IoU between two shapes $\gG$ and $\rR$, represented as binary occupancy maps, is commonly defined as
\begin{align}
	\text{IoU}(\gG,\rR) &= \frac{\abs{\gG\cap \rR}}{\abs{\gG \cup \rR}}.
\end{align}
In our evaluation protocol, we compare shapes $A,B$ at a resolution of $128^3$ binary cells (voxels).

\subsection{Chamfer Distance (CD)}

The Chamfer Distance (CD) between the ground truth shape $\gG$ and the reconstructed shape $\rR$ (both represented as point clouds) is defined as
\begin{align}
\begin{split}
\text{CD}(\gG,\rR) = \frac{1}{\abs{\rR}}&\sum_{r \in \rR}\min_{g \in \gG} \norm{r-g}_2\\ + \frac{1}{\abs{\gG}}&\sum_{g \in \gG}\min_{r \in \rR} \norm{g-r}_2.
\end{split}
\end{align}
\subsection{F-score}

Here we provide a full definition of the F-score measure.
Consider a ground truth shape $\gG$ and a reconstructed shape $\rR$ both represented as point clouds.
For every point $r \in \rR$ its distance to $\gG$ is calculated as
\begin{equation*}
e_{r}=\min_{g \in \gG} \norm{r-g}_2.
\end{equation*}
Subsequently, we calculate the percentage of points reconstructed better than a certain threshold $d$ which results in the \textit{precision} value
\begin{equation*}
    P(d) = \frac{100}{\abs{\rR}}\sum_{r \in \rR}[e_r < d].
\end{equation*}
The same procedure is repeated in the opposite direction to produce the \textit{recall} value
\begin{equation*}
    e_{g} =\min_{r \in \rR} \norm{g-r}_2,\qquad
    R(d)  = \frac{100}{\abs{\gG}}\sum_{g \in \gG}[e_g < d].
\end{equation*}
The final F-score is given by the harmonic mean of the precision and recall values
\begin{equation}
    F(d) = \frac{2P(d)R(d)}{P(d) + R(d)}.
\end{equation}
In practice, we set $d$ as a fraction of the side length of the reconstructed volume (\eg,{} $1\%$).

To evaluate a method using the F-score, we convert each shape prediction to a mesh representation, from which we evenly sample 10K points from the surface. We show how predictions by different methods compare in terms of their visual quality, precision and recall for a qualitative example in Fig.~\ref{fig:precision_recall_qual}.
OGN~\cite{tatarchenko17}, Matryoshka~\cite{richter18} and the clustering baseline completely miss parts of the plane, resulting in high precision but comparably low recall.
AtlasNet~\cite{groueix18} reconstructs a complete shape, but misplaces individual parts, resulting in both low precision and low recall.
The retrieval baseline finds a reasonably similar model, leading to comparably high precision and recall values.

\section{Quantitative results}

In Tab.~\ref{tbl:results} we provide the exact F-score values at 1\% threshold in the viewer-centered mode.

\begin{table*}[htb!]
\small
\begin{center}
\setlength{\tabcolsep}{2mm}
\begin{tabular}{@{}l c c c c c}
\toprule
 & \textbf{AtlasNet} & \textbf{OGN} & \textbf{Matryoshka} & \textbf{Retrieval} & \textbf{Oracle NN} \\
\midrule
airplane & 0.39 & 0.26 & 0.33 & 0.37 & 0.45 \\
ashcan & 0.18 & 0.23 & 0.26 & 0.21 & 0.24 \\
bag & 0.16 & 0.14 & 0.18 & 0.13 & 0.15 \\
basket & 0.19 & 0.16 & 0.21 & 0.15 & 0.15 \\
bathtub & 0.25 & 0.13 & 0.26 & 0.22 & 0.26 \\
bed & 0.19 & 0.12 & 0.18 & 0.15 & 0.17 \\
bench & 0.34 & 0.09 & 0.32 & 0.3 & 0.34 \\
birdhouse & 0.17 & 0.13 & 0.18 & 0.15 & 0.15 \\
bookshelf & 0.24 & 0.18 & 0.25 & 0.2 & 0.2 \\
bottle & 0.34 & 0.54 & 0.45 & 0.46 & 0.55 \\
bowl & 0.22 & 0.18 & 0.24 & 0.2 & 0.25 \\
bus & 0.35 & 0.38 & 0.41 & 0.36 & 0.44 \\
cabinet & 0.25 & 0.29 & 0.33 & 0.23 & 0.27 \\
camera & 0.13 & 0.08 & 0.12 & 0.11 & 0.12 \\
can & 0.23 & 0.46 & 0.44 & 0.36 & 0.44 \\
cap & 0.18 & 0.02 & 0.15 & 0.19 & 0.25 \\
car & 0.3 & 0.37 & 0.38 & 0.33 & 0.39 \\
cellular & 0.34 & 0.45 & 0.47 & 0.41 & 0.5 \\
chair & 0.25 & 0.15 & 0.27 & 0.2 & 0.23 \\
clock & 0.24 & 0.21 & 0.25 & 0.22 & 0.27 \\
dishwasher & 0.2 & 0.29 & 0.31 & 0.22 & 0.26 \\
display & 0.22 & 0.15 & 0.23 & 0.19 & 0.24 \\
earphone & 0.14 & 0.07 & 0.11 & 0.11 & 0.13 \\
faucet & 0.19 & 0.06 & 0.13 & 0.14 & 0.2 \\
file & 0.22 & 0.33 & 0.36 & 0.24 & 0.25 \\
guitar & 0.45 & 0.35 & 0.36 & 0.41 & 0.58 \\
helmet & 0.1 & 0.06 & 0.09 & 0.08 & 0.12 \\
jar & 0.21 & 0.22 & 0.25 & 0.19 & 0.22 \\
keyboard & 0.36 & 0.25 & 0.37 & 0.35 & 0.49 \\
knife & 0.46 & 0.26 & 0.21 & 0.37 & 0.54 \\
lamp & 0.26 & 0.13 & 0.2 & 0.21 & 0.27 \\
laptop & 0.29 & 0.21 & 0.33 & 0.26 & 0.33 \\
loudspeaker & 0.2 & 0.26 & 0.27 & 0.19 & 0.23 \\
mailbox & 0.21 & 0.2 & 0.23 & 0.2 & 0.19 \\
microphone & 0.23 & 0.22 & 0.19 & 0.18 & 0.21 \\
microwave & 0.23 & 0.36 & 0.35 & 0.22 & 0.25 \\
motorcycle & 0.27 & 0.12 & 0.22 & 0.24 & 0.28 \\
mug & 0.13 & 0.11 & 0.15 & 0.11 & 0.17 \\
piano & 0.17 & 0.11 & 0.16 & 0.14 & 0.17 \\
pillow & 0.19 & 0.14 & 0.17 & 0.18 & 0.3 \\
pistol & 0.29 & 0.22 & 0.23 & 0.25 & 0.3 \\
pot & 0.19 & 0.15 & 0.19 & 0.14 & 0.16 \\
printer & 0.13 & 0.11 & 0.13 & 0.11 & 0.14 \\
remote & 0.3 & 0.33 & 0.31 & 0.31 & 0.37 \\
rifle & 0.43 & 0.28 & 0.3 & 0.36 & 0.48 \\
rocket & 0.34 & 0.2 & 0.23 & 0.26 & 0.32 \\
skateboard & 0.39 & 0.11 & 0.39 & 0.35 & 0.47 \\
sofa & 0.24 & 0.23 & 0.27 & 0.21 & 0.27 \\
stove & 0.2 & 0.19 & 0.24 & 0.18 & 0.19 \\
table & 0.31 & 0.24 & 0.34 & 0.26 & 0.34 \\
telephone & 0.33 & 0.42 & 0.45 & 0.4 & 0.5 \\
tower & 0.24 & 0.2 & 0.25 & 0.25 & 0.25 \\
train & 0.34 & 0.29 & 0.3 & 0.32 & 0.38 \\
vessel & 0.28 & 0.19 & 0.22 & 0.23 & 0.29 \\
washer & 0.2 & 0.31 & 0.31 & 0.21 & 0.25 \\
\bottomrule
\end{tabular}
\caption{F-score evaluation (@1\%) in the viewer-centered mode.}
\label{tbl:results}
\vspace{-2mm}
\end{center}
\end{table*}

\section{Qualitative examples}
In addition to the qualitative examples for a selection of classes in the main paper, we show a randomly sampled qualitative example for each class of the ShapeNet dataset in Fig.~\ref{fig:qualitative_0}. As in the main paper, we show, from left to right: input image, ground truth shape, and predictions from AtlasNet~\cite{groueix18}, OGN~\cite{tatarchenko17}, Matryoshka~\cite{richter18},  our clustering baseline, our retrieval baseline, and an Oracle Nearest Neighbor.
Numbers in the bottom left of each prediction indicate the IoU (dark gray) and the F-score at a $1\%$ threshold (bold), respectively.

\section{Statistical evaluation}

In the main paper we showed within-class IoU histograms for a selection of three classes.
We visualize such histograms for all 55 classes in Fig.~\ref{fig:iou_histograms1}.

We performed the Kolmogorov-Smirnov test on the within-class distributions for each ShapeNet class and each pairing of methods. The null hypothesis assumes that two distributions exhibit no statistically significant difference. We plot the \textit{p}-values for each test result in Fig.~\ref{fig:pvalue_iou}. The color of each cell indicates whether the null hypothesis can be rejected (orange) or not (green).
Aggregated results can be found in Fig.~6~(right) in the main paper.

\newpage

\begin{figure*}[ht]
\centering
\begin{tabular}{@{}cc@{\hspace{8mm}}c@{\hspace{5mm}}c@{}}
& Input & \multicolumn{2}{c}{Ground truth} \\
& \includegraphics[clip, trim=0 0 0 1cm, height=2.5cm]{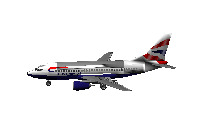} &
\multicolumn{2}{c}{\includegraphics[clip, trim=0 0 0 1cm,height=2.5cm]{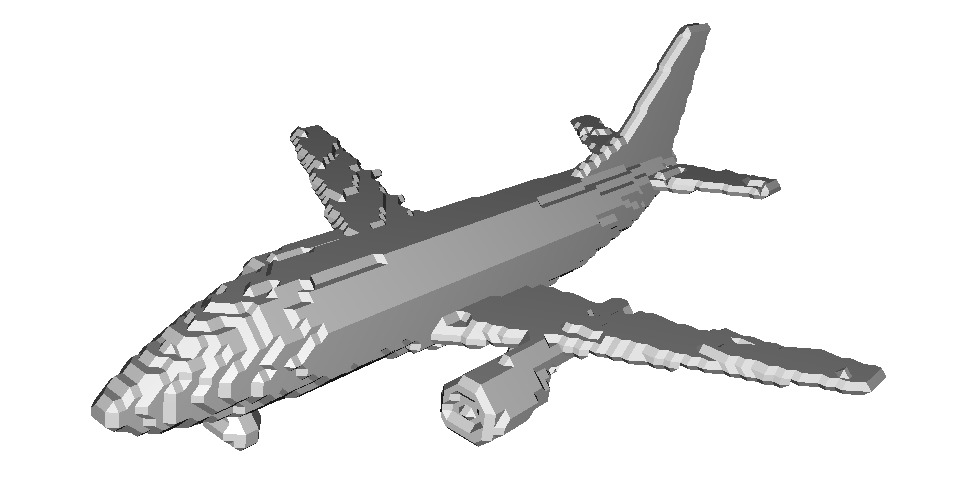}}\\[0.5em]
& Prediction & Precision & Recall \\
\rotc{\footnotesize AtlasNet} & 
\raisebox{-0.5\height}{\includegraphics[clip,trim=2cm 0 0 0,height=2.5cm]{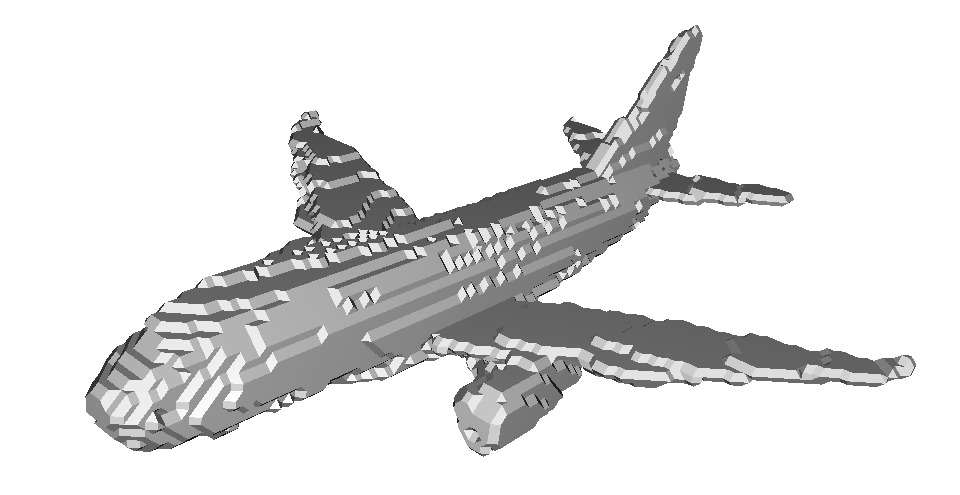}} &
\raisebox{-0.5\height}{\includegraphics[clip,trim=2cm 0 0 0,height=2.5cm]{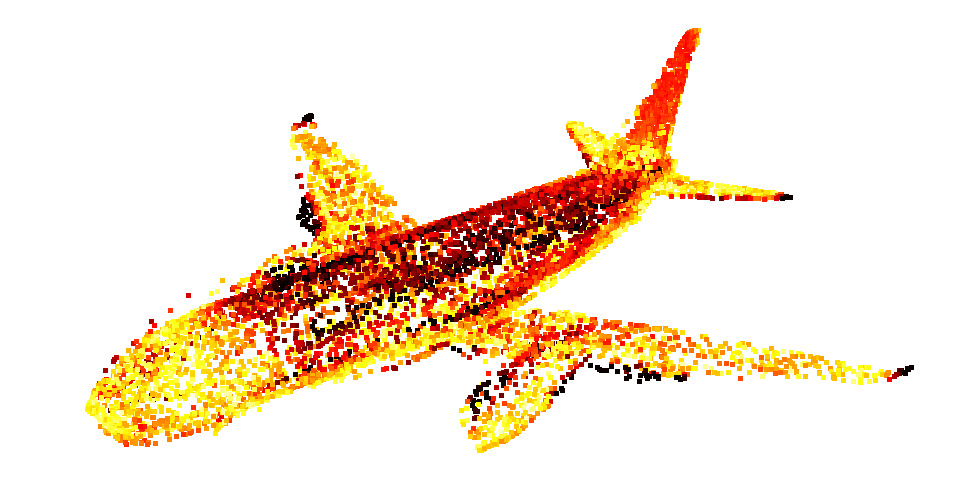}} &
\raisebox{-0.5\height}{\includegraphics[clip,trim=2cm 0 0 0,height=2.5cm]{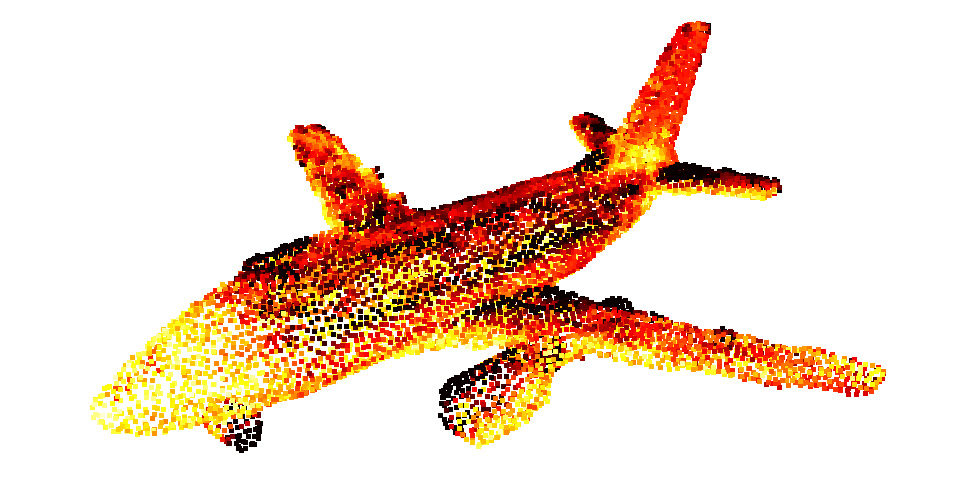}} \\
& \footnotesize IoU = 0.68, F = 0.52 & \footnotesize P = 0.57 & \footnotesize R = 0.47 \\
\rotc{\footnotesize OGN} & 
\raisebox{-0.5\height}{\includegraphics[clip,trim=2cm 0 0 0,height=2.5cm]{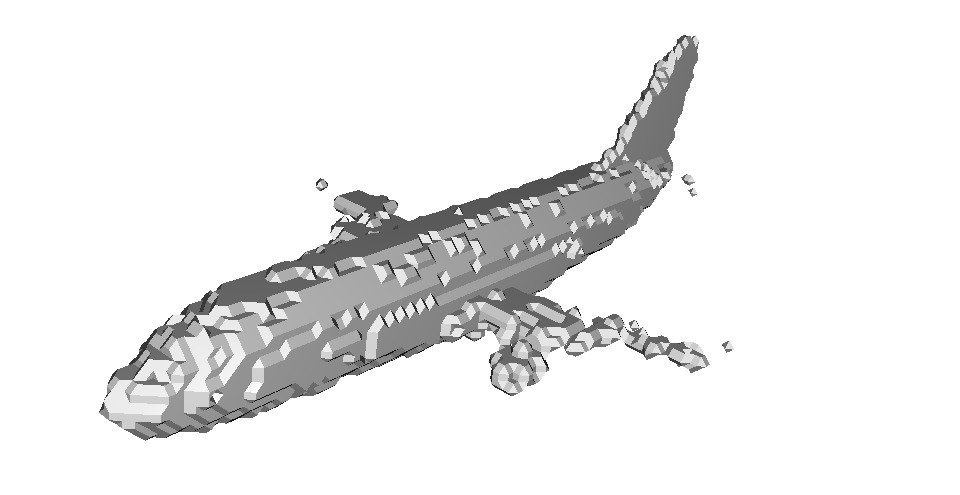}} &
\raisebox{-0.5\height}{\includegraphics[clip,trim=2cm 0 0 0,height=2.5cm]{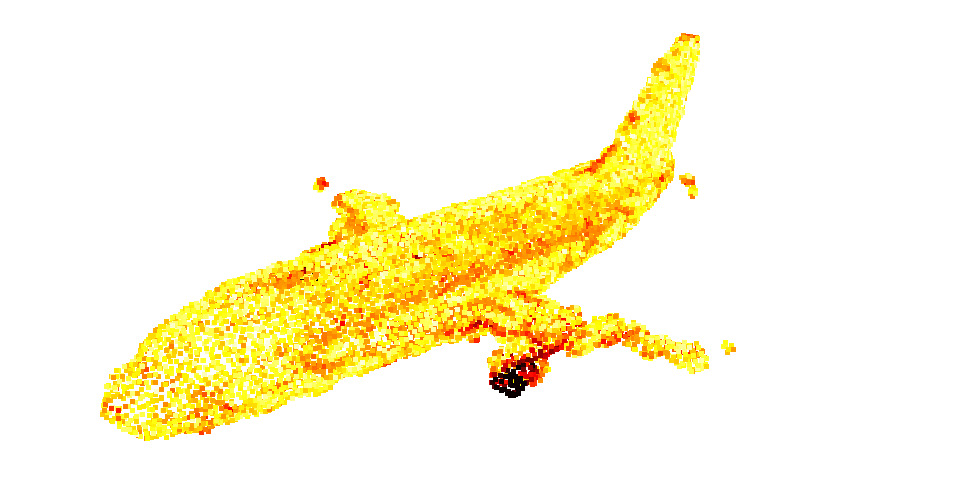}} &
\raisebox{-0.5\height}{\includegraphics[clip,trim=2cm 0 0 0,height=2.5cm]{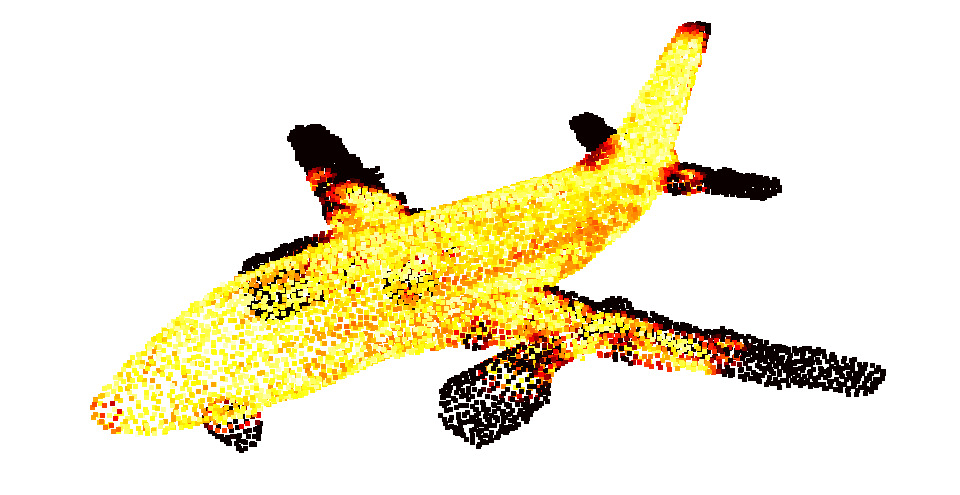}} \\
& \footnotesize IoU = 0.73, F = 0.76 & \footnotesize P = 0.94 & \footnotesize R = 0.64 \\
\rotc{\footnotesize Matryoshka} & 
\raisebox{-0.5\height}{\includegraphics[clip,trim=2cm 0 0 0,height=2.5cm]{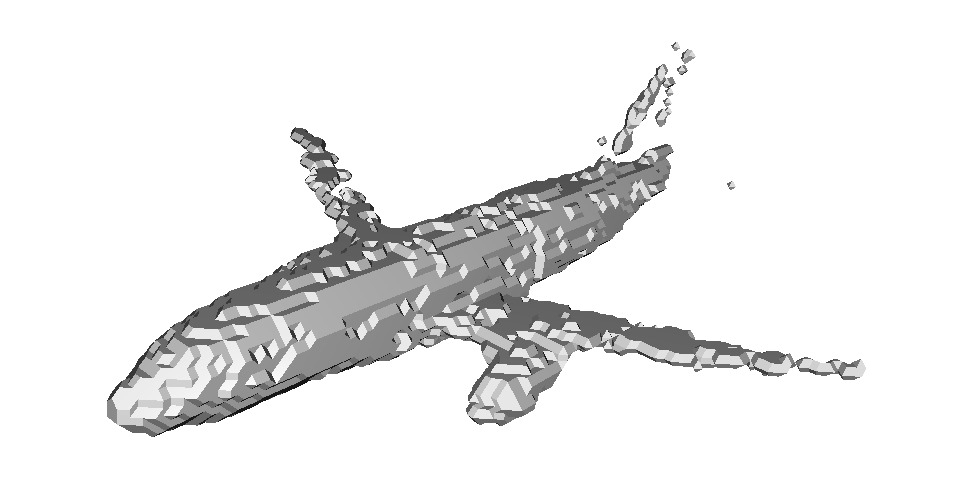}} & 
\raisebox{-0.5\height}{\includegraphics[clip,trim=2cm 0 0 0,height=2.5cm]{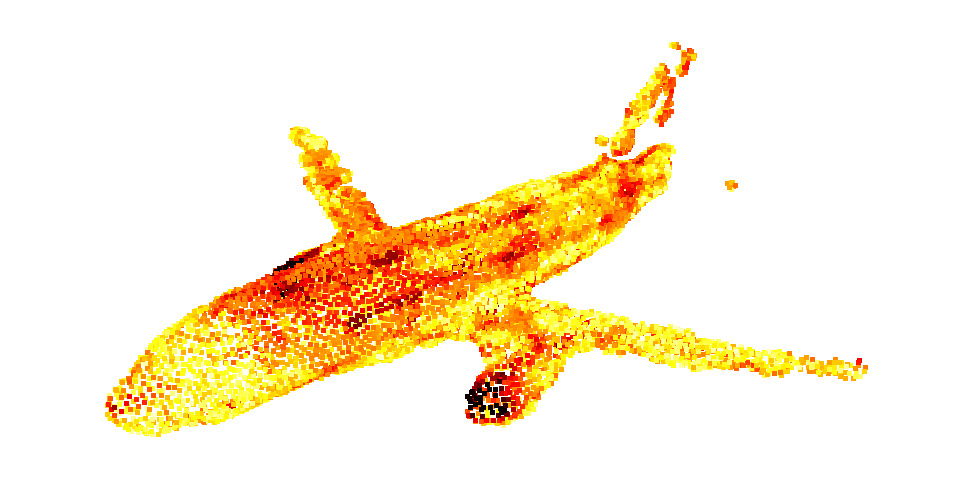}} &
\raisebox{-0.5\height}{\includegraphics[clip,trim=2cm 0 0 0,height=2.5cm]{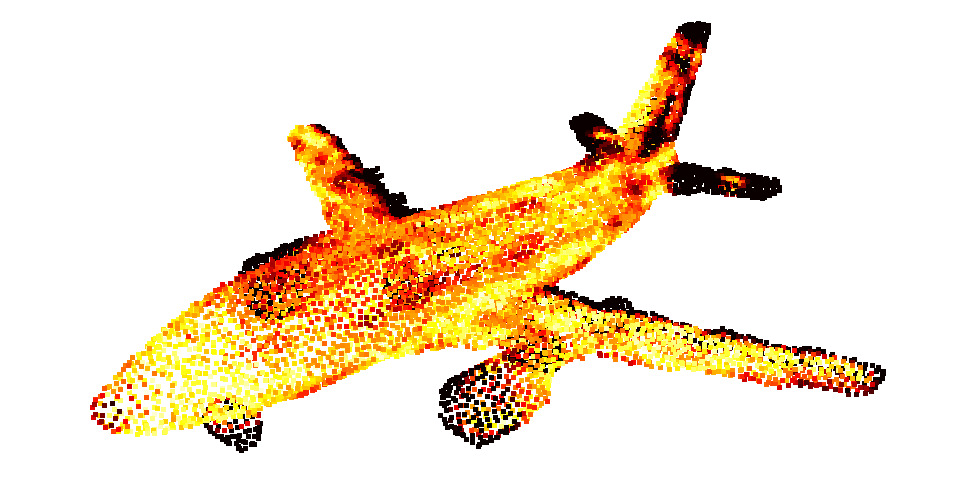}} \\
\end{tabular}
\caption{Exemplary predictions of different methods compared by visual quality, precision and recall. Colors encode the point-to-surface distance between shapes, normalized by the side length of the reconstructed volume.}
\label{fig:precision_recall_qual}
\end{figure*}
\begin{figure*}
\begin{tabular}{@{}*{8}{c@{\hspace{0.2mm}}}c@{}}
& \footnotesize Input & 
\footnotesize Ground truth & 
\footnotesize AtlasNet & 
\footnotesize OGN & 
\footnotesize Matryoshka & 
\footnotesize Clustering & 
\footnotesize Retrieval & 
\footnotesize Oracle NN\\
\end{tabular}
\caption{Qualitative results for all classes of ShapeNet.
Numbers in the bottom right corner of each sample indicate IoU~(dark gray) and F-score~(bold), respectively.}
\label{fig:qualitative_0}
\end{figure*}
\begin{figure*}
\ContinuedFloat
\begin{tabular}{@{}*{8}{c@{\hspace{0.2mm}}}c@{}}
& \footnotesize Input & 
\footnotesize Ground truth & 
\footnotesize AtlasNet & 
\footnotesize OGN & 
\footnotesize Matryoshka & 
\footnotesize Clustering & 
\footnotesize Retrieval & 
\footnotesize Oracle NN\\
\rotc{\footnotesize bowl} &
\raisebox{-0.5\height}{\includegraphics[width=0.10\linewidth]{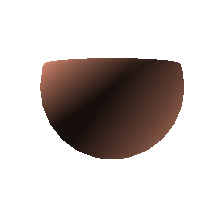}} &
\raisebox{-0.5\height}{\includegraphics[clip,trim=0 0 0 0,width=0.12\linewidth]{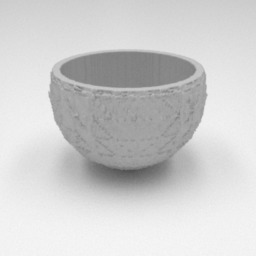}} &
\raisebox{-0.5\height}{\begin{overpic}[clip,trim=0 0 0 0,width=0.12\linewidth]{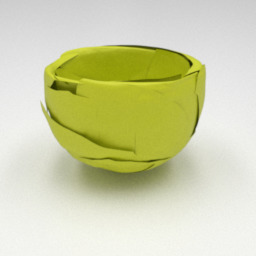}
	\put (77,14) {\color{black!70} \scriptsize 0.48}
	\put (77,4) {\scriptsize \textbf{0.22}}
\end{overpic}}&
\raisebox{-0.5\height}{\begin{overpic}[clip,trim=0 0 0 0,width=0.12\linewidth]{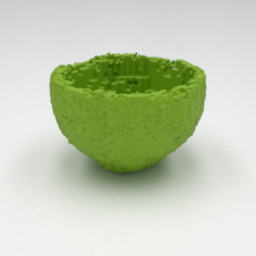}
	\put (77,14) {\color{black!70} \scriptsize 0.50}
	\put (77,4) {\scriptsize \textbf{0.21}}
\end{overpic}} &
\raisebox{-0.5\height}{\begin{overpic}[clip,trim=0 0 0 0,width=0.12\linewidth]{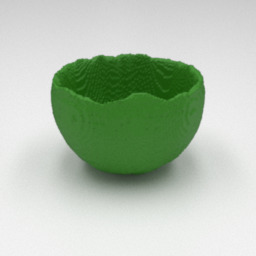}
	\put (77,14) {\color{black!70} \scriptsize 0.67}
	\put (77,4) {\scriptsize \textbf{0.32}}
\end{overpic}} &
\raisebox{-0.5\height}{\begin{overpic}[clip,trim=0 0 0 0,width=0.12\linewidth]{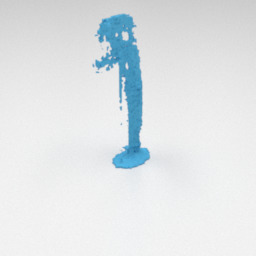}
	\put (77,14) {\color{black!70} \scriptsize 0.00}
	\put (77,4) {\scriptsize \textbf{0.01}}
\end{overpic}} &
\raisebox{-0.5\height}{\begin{overpic}[clip,trim=0 0 0 0,width=0.12\linewidth]{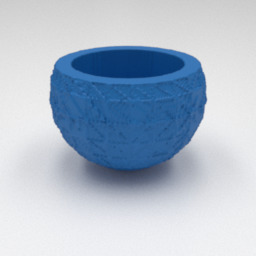}
	\put (77,14) {\color{black!70} \scriptsize 0.46}
	\put (77,4) {\scriptsize \textbf{0.13}}
\end{overpic}} &
\raisebox{-0.5\height}{\begin{overpic}[clip,trim=0 0 0 0,width=0.12\linewidth]{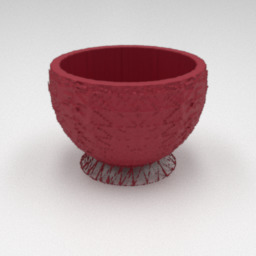}
	\put (77,14) {\color{black!70} \scriptsize 0.68}
	\put (77,4) {\scriptsize \textbf{0.30}}
\end{overpic}}\\
\end{tabular}
\caption{Qualitative results for all classes of ShapeNet (continued).}
\label{fig:qualitative_1}
\end{figure*}
\begin{figure*}
\ContinuedFloat
\begin{tabular}{@{}*{8}{c@{\hspace{0.2mm}}}c@{}}
& \footnotesize Input & 
\footnotesize Ground truth & 
\footnotesize AtlasNet & 
\footnotesize OGN & 
\footnotesize Matryoshka & 
\footnotesize Clustering & 
\footnotesize Retrieval & 
\footnotesize Oracle NN\\
\end{tabular}
\caption{Qualitative results for all classes of ShapeNet (continued).}
\label{fig:qualitative_2}
\end{figure*}
\begin{figure*}
\ContinuedFloat
\begin{tabular}{@{}*{8}{c@{\hspace{0.2mm}}}c@{}}
& \footnotesize Input & 
\footnotesize Ground truth & 
\footnotesize AtlasNet & 
\footnotesize OGN & 
\footnotesize Matryoshka & 
\footnotesize Clustering & 
\footnotesize Retrieval & 
\footnotesize Oracle NN\\
\rotc{\footnotesize lamp} &
\raisebox{-0.5\height}{\includegraphics[width=0.10\linewidth]{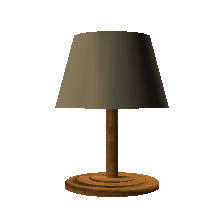}} &
\raisebox{-0.5\height}{\includegraphics[clip,trim=0 0 0 0,width=0.12\linewidth]{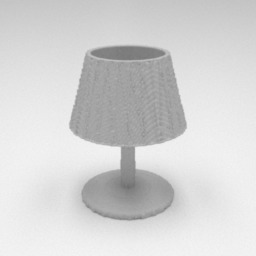}} &
\raisebox{-0.5\height}{\begin{overpic}[clip,trim=0 0 0 0,width=0.12\linewidth]{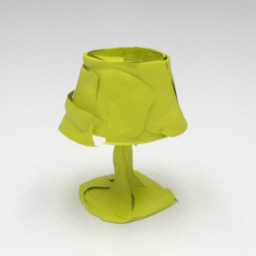}
	\put (77,14) {\color{black!70} \scriptsize 0.49}
	\put (77,4) {\scriptsize \textbf{0.18}}
\end{overpic}}&
\raisebox{-0.5\height}{\begin{overpic}[clip,trim=0 0 0 0,width=0.12\linewidth]{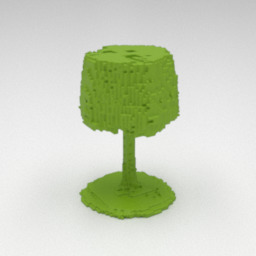}
	\put (77,14) {\color{black!70} \scriptsize 0.42}
	\put (77,4) {\scriptsize \textbf{0.30}}
\end{overpic}} &
\raisebox{-0.5\height}{\begin{overpic}[clip,trim=0 0 0 0,width=0.12\linewidth]{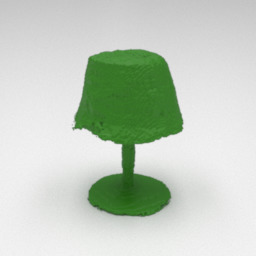}
	\put (77,14) {\color{black!70} \scriptsize 0.35}
	\put (77,4) {\scriptsize \textbf{0.19}}
\end{overpic}} &
\raisebox{-0.5\height}{\begin{overpic}[clip,trim=0 0 0 0,width=0.12\linewidth]{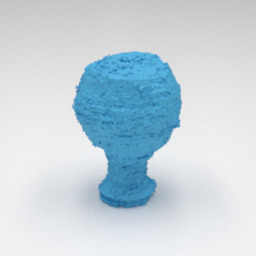}
	\put (77,14) {\color{black!70} \scriptsize 0.31}
	\put (77,4) {\scriptsize \textbf{0.19}}
\end{overpic}} &
\raisebox{-0.5\height}{\begin{overpic}[clip,trim=0 0 0 0,width=0.12\linewidth]{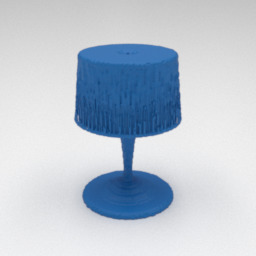}
	\put (77,14) {\color{black!70} \scriptsize 0.42}
	\put (77,4) {\scriptsize \textbf{0.29}}
\end{overpic}} &
\raisebox{-0.5\height}{\begin{overpic}[clip,trim=0 0 0 0,width=0.12\linewidth]{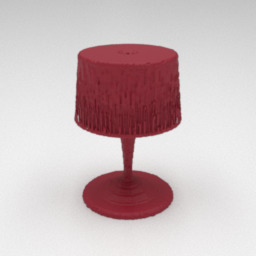}
	\put (77,14) {\color{black!70} \scriptsize 0.42}
	\put (77,4) {\scriptsize \textbf{0.29}}
\end{overpic}}\\
\end{tabular}
\caption{Qualitative results for all classes of ShapeNet (continued).}
\label{fig:qualitative_3}
\end{figure*}
\begin{figure*}
\ContinuedFloat
\begin{tabular}{@{}*{8}{c@{\hspace{0.2mm}}}c@{}}
& \footnotesize Input & 
\footnotesize Ground truth & 
\footnotesize AtlasNet & 
\footnotesize OGN & 
\footnotesize Matryoshka & 
\footnotesize Clustering & 
\footnotesize Retrieval & 
\footnotesize Oracle NN\\
\end{tabular}
\caption{Qualitative results for all classes of ShapeNet (continued).}
\label{fig:qualitative_4}
\end{figure*}
\begin{figure*}
\ContinuedFloat
\begin{tabular}{@{}*{8}{c@{\hspace{0.2mm}}}c@{}}
& \footnotesize Input & 
\footnotesize Ground truth & 
\footnotesize AtlasNet & 
\footnotesize OGN & 
\footnotesize Matryoshka & 
\footnotesize Clustering & 
\footnotesize Retrieval & 
\footnotesize Oracle NN\\
\end{tabular}
\caption{Qualitative results for all classes of ShapeNet (continued).}
\label{fig:qualitative_5}
\end{figure*}

\begin{figure*}[ht]
\centering
\includegraphics[width=17.5cm]{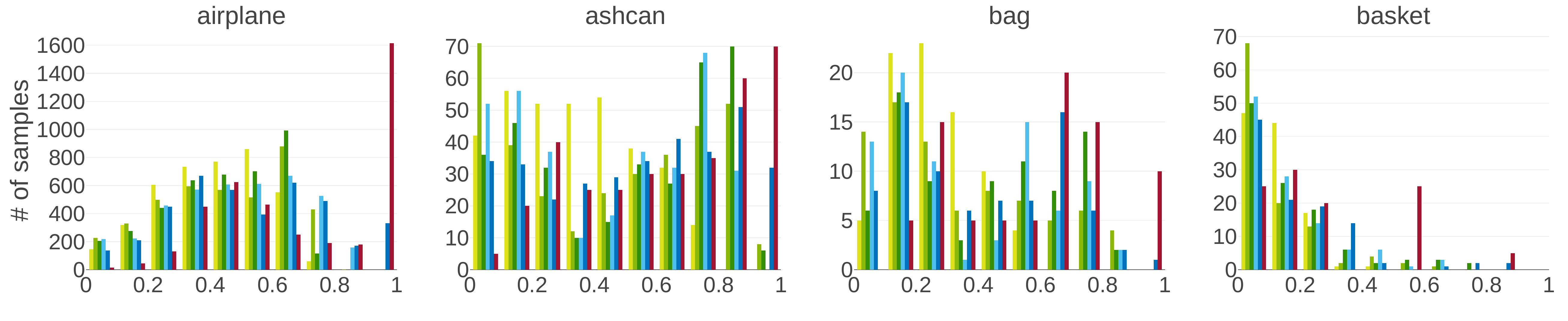}
\includegraphics[width=17.5cm]{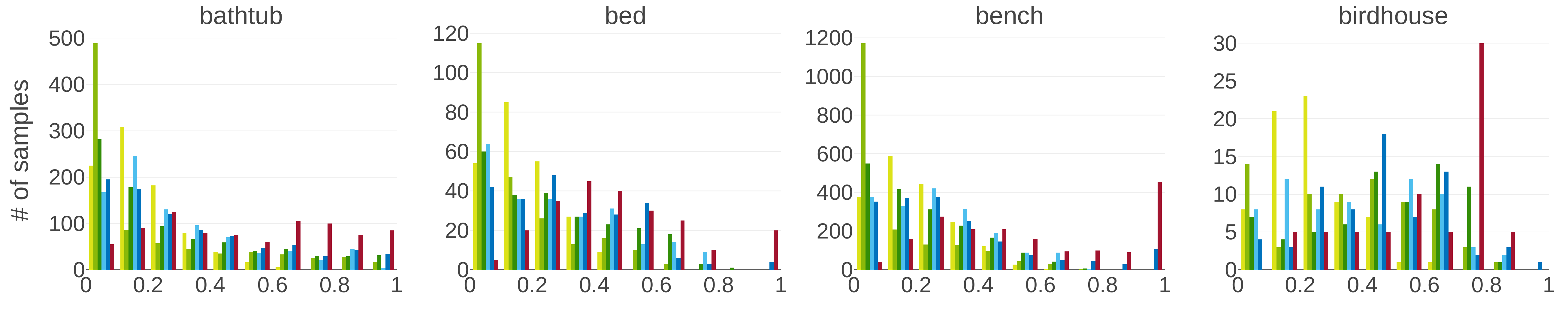}
\includegraphics[width=17.5cm]{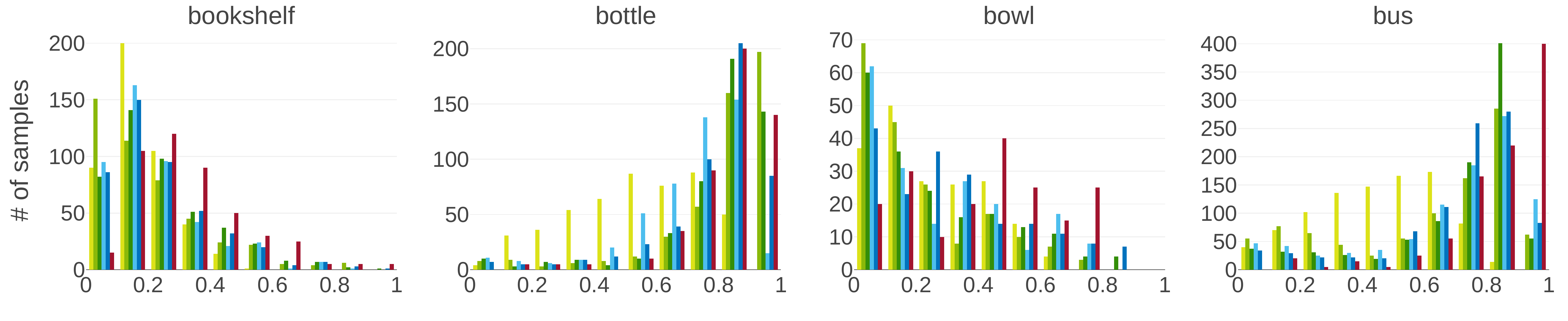}
\includegraphics[width=17.5cm]{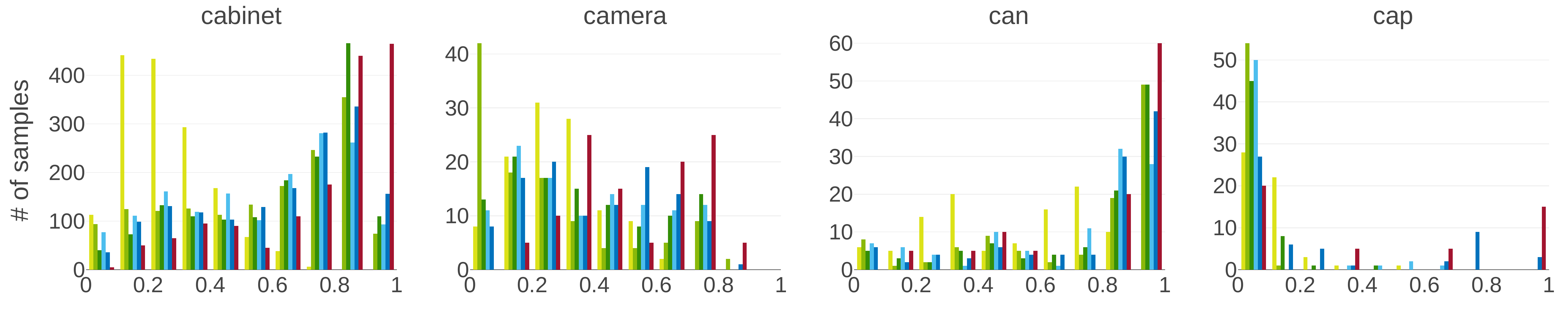}
\includegraphics[width=17.5cm]{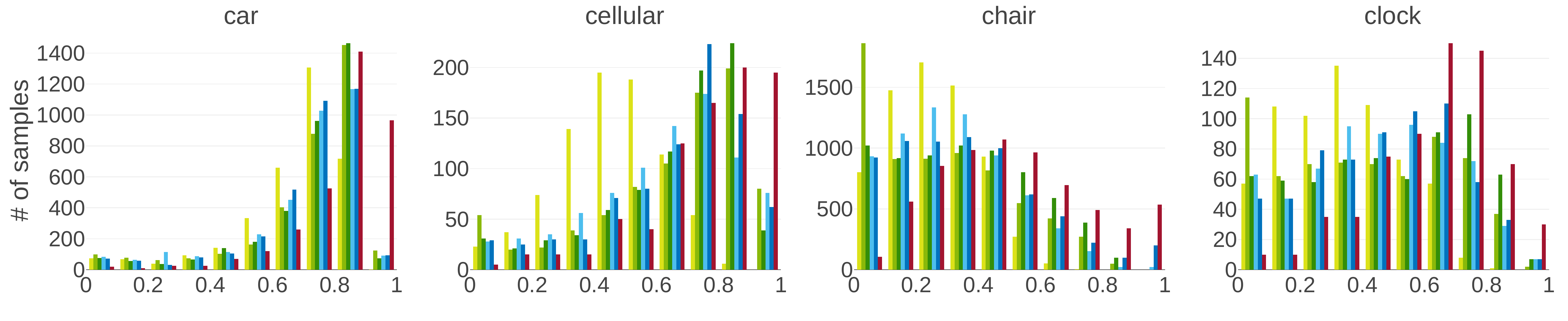}
\includegraphics[width=17.5cm]{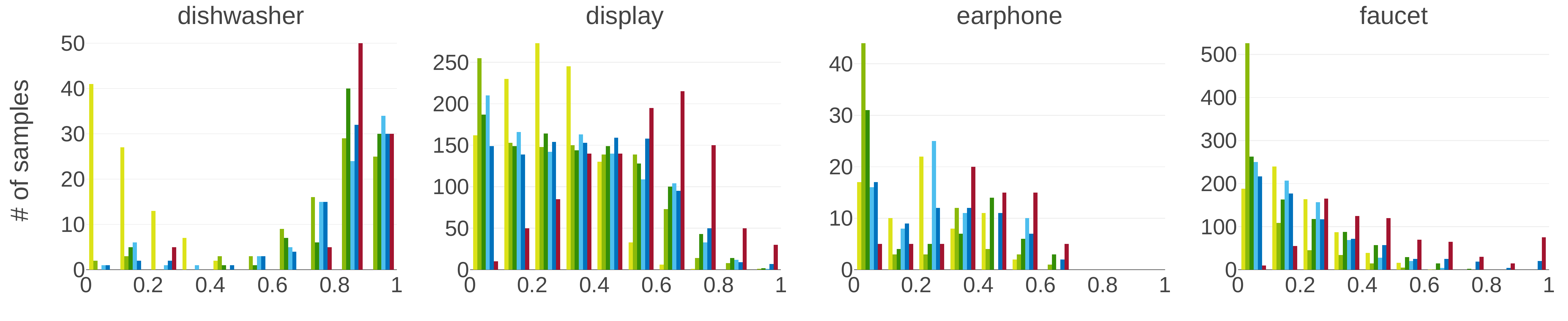}
\caption{Distribution of within-class reconstruction performance for all ShapeNet classes, measured by IoU.}
\label{fig:iou_histograms1}
\end{figure*}
\begin{figure*}[ht]\ContinuedFloat
\centering
\includegraphics[width=17.5cm]{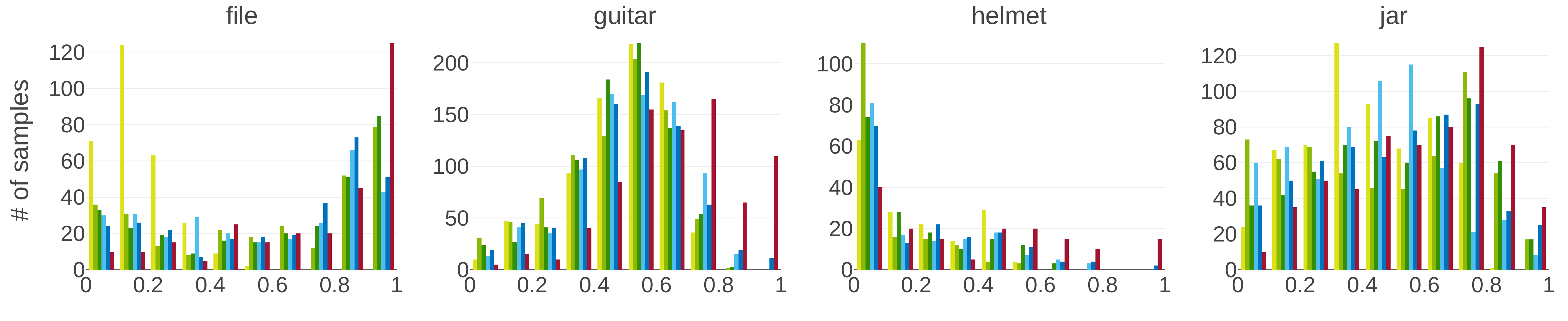}
\includegraphics[width=17.5cm]{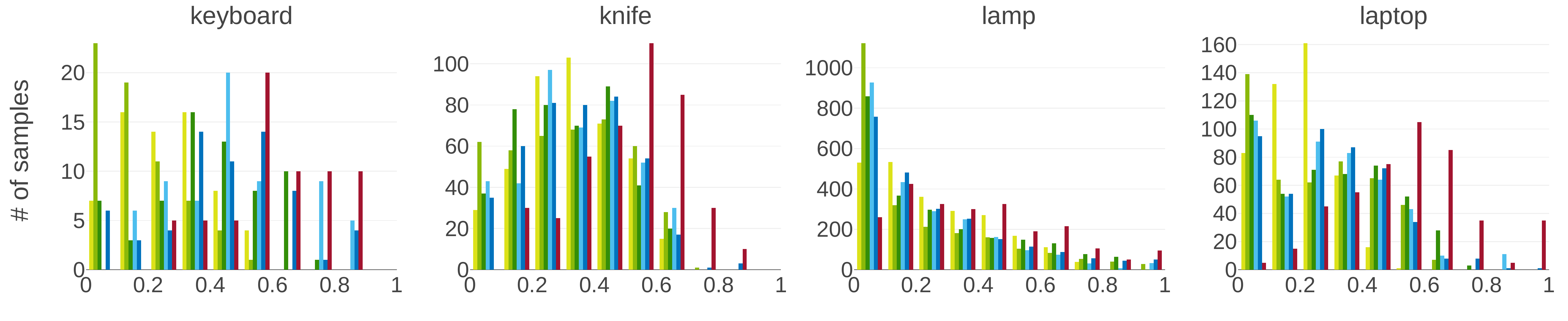}
\includegraphics[width=17.5cm]{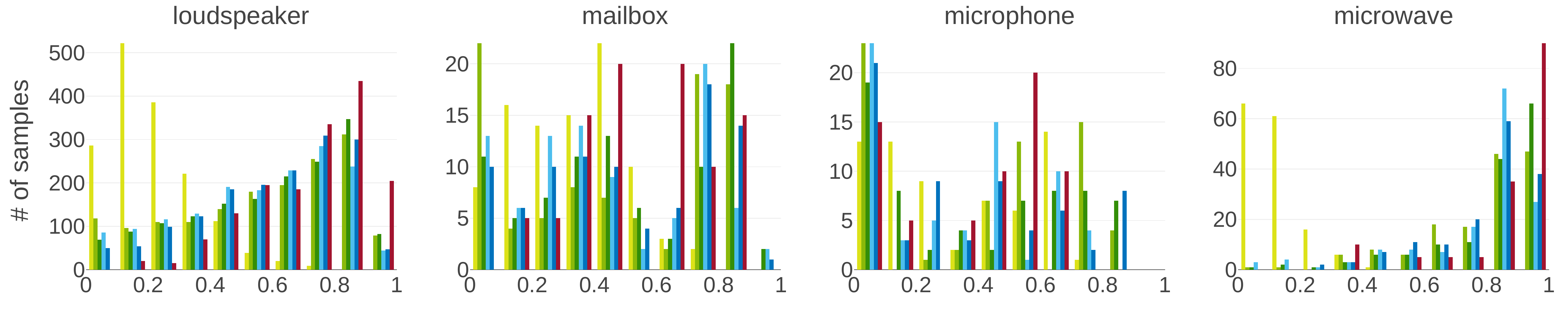}
\includegraphics[width=17.5cm]{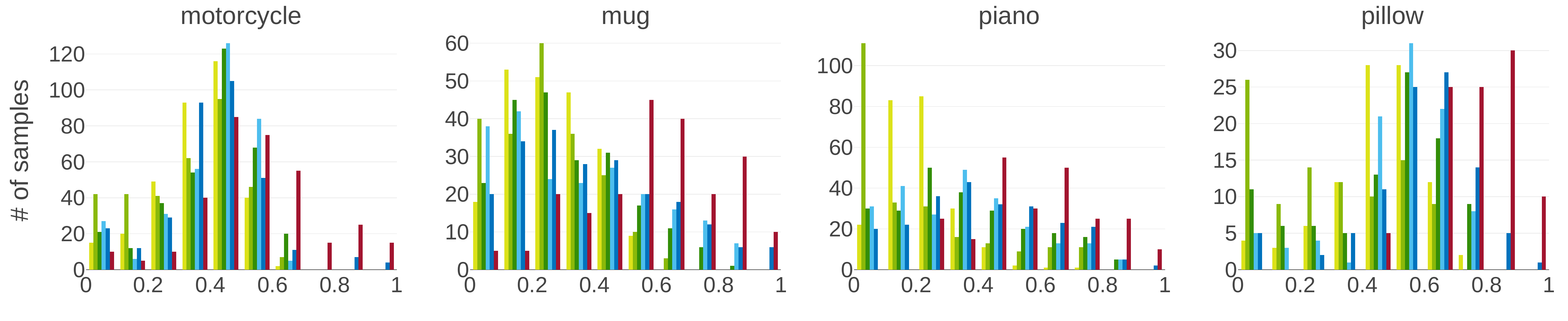}
\includegraphics[width=17.5cm]{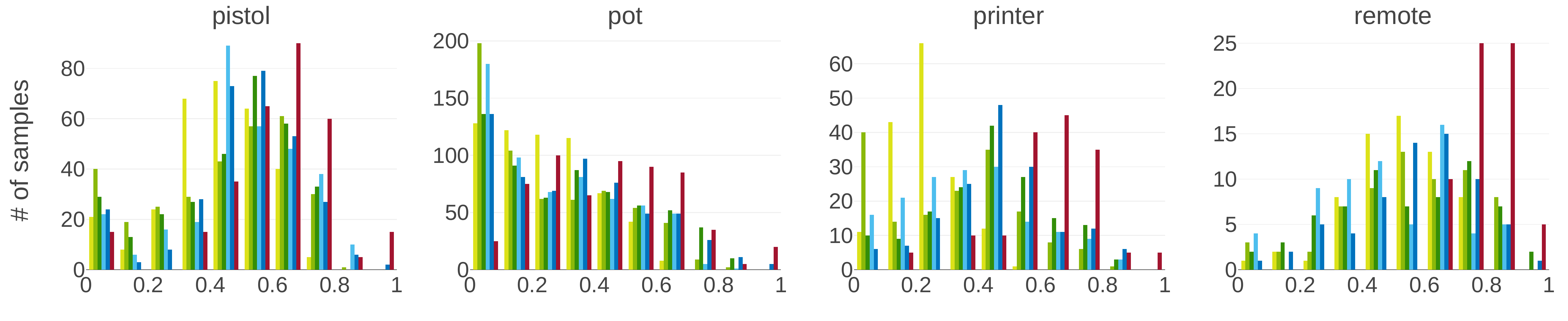}
\includegraphics[width=17.5cm]{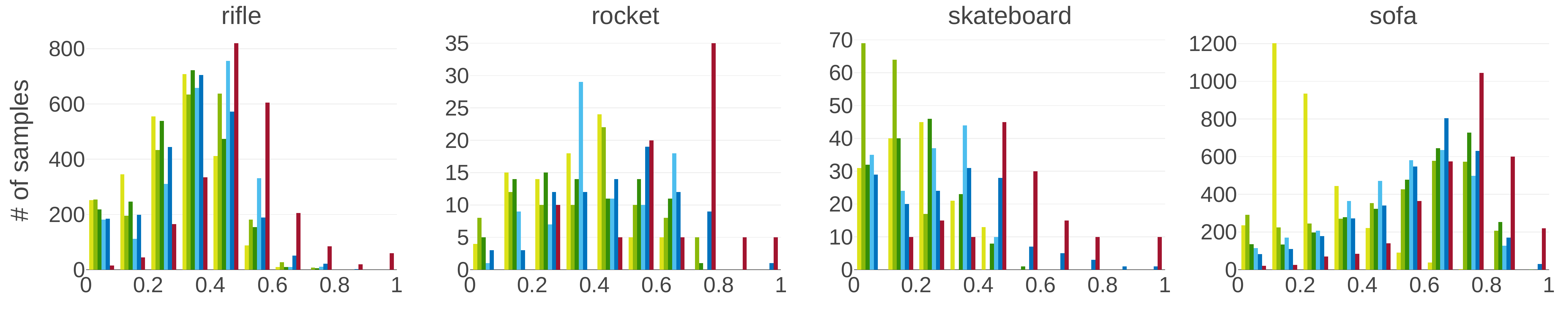}
\caption{Distribution of within-class reconstruction performance for all ShapeNet classes, measured by IoU (continued).}
\label{fig:iou_histograms2}
\end{figure*}
\begin{figure*}[ht]\ContinuedFloat
\centering
\includegraphics[width=17.5cm]{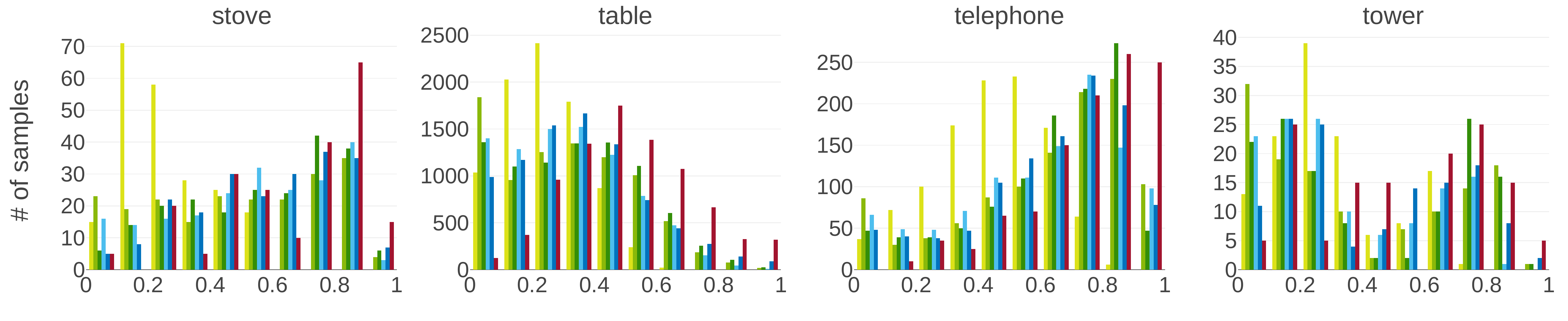}
\includegraphics[width=17.5cm]{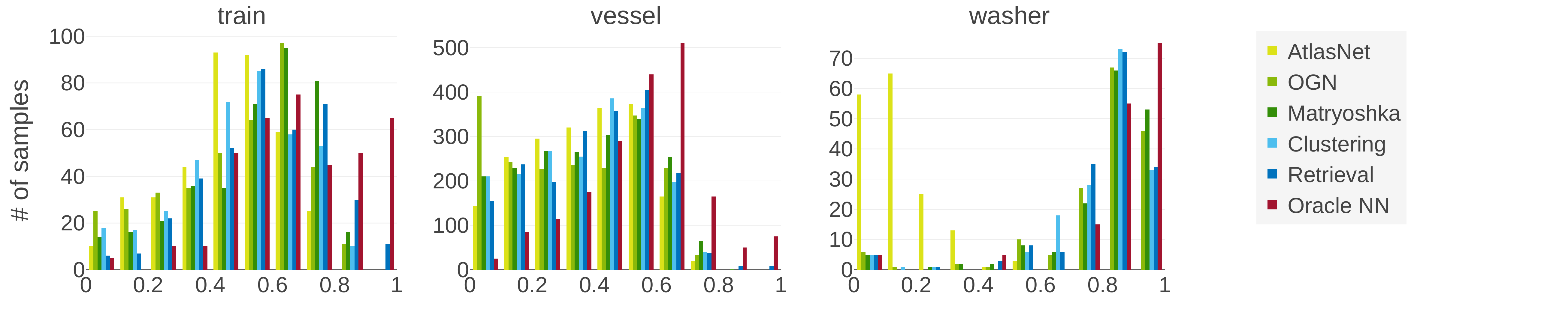}
\caption{Distribution of within-class reconstruction performance for all ShapeNet classes, measured by IoU (continued).}
\label{fig:iou_histograms3}
\end{figure*}
\begin{figure*}[ht]
\centering
\includegraphics[width=17.5cm]{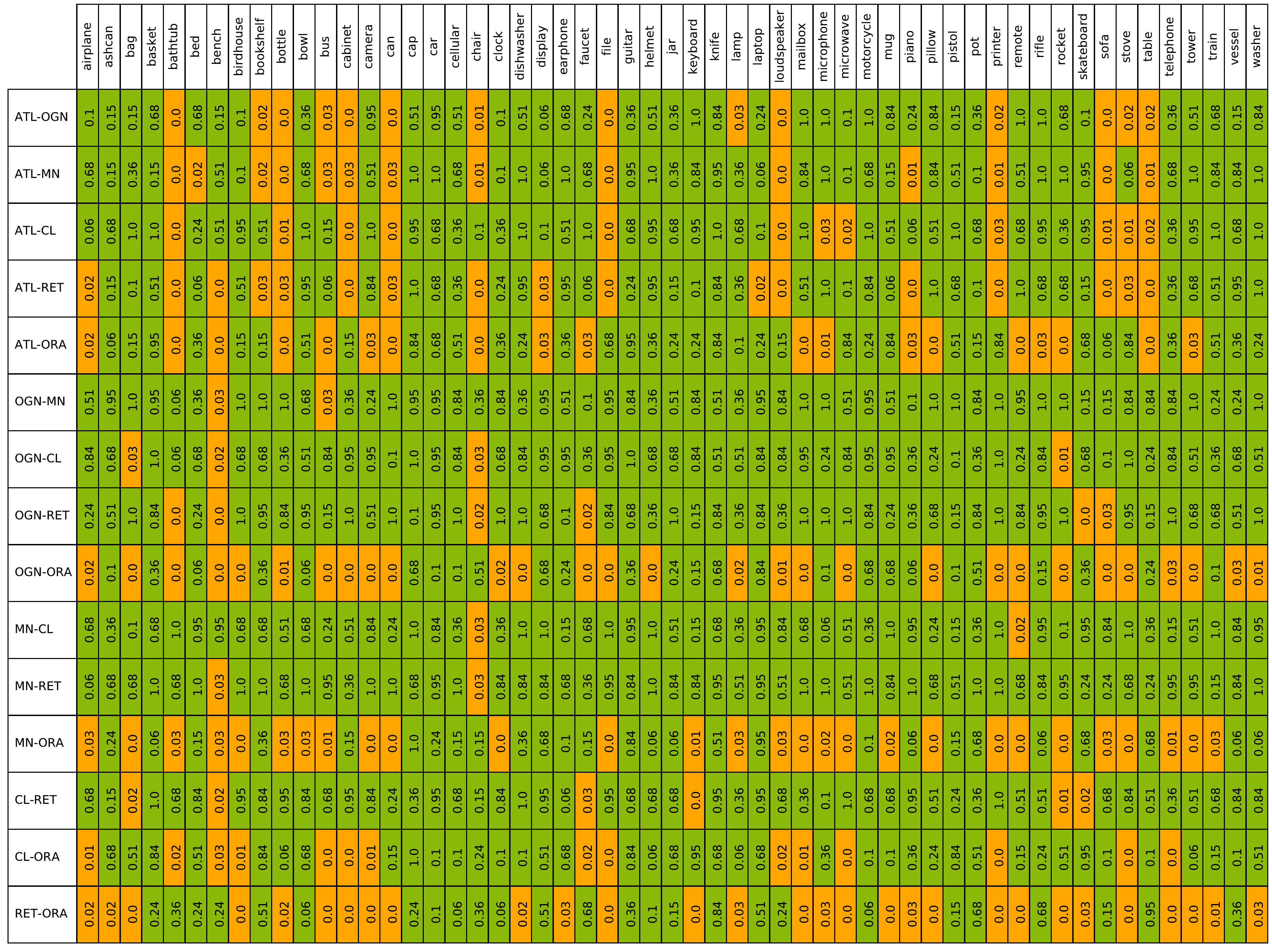}
\caption{P-values of the pairwise Kolmogorov-Smirnov test on per-class IoU performance histograms. The null-hypotheses of two distributions being the same can be rejected for $p<0.05$ (orange) and cannot be rejected for $p>0.05$ (green).}
\label{fig:pvalue_iou}
\end{figure*}

\end{document}